\documentclass[a4paper, reqno, 12pt]{amsart}

\usepackage[utf8]{inputenc}
\usepackage{graphicx}
\usepackage{booktabs}
\usepackage{array}
\usepackage{hyperref}
\usepackage{url}
\usepackage{xcolor}
\usepackage{fancyhdr}
\usepackage{geometry}
\usepackage{amssymb} 
\usepackage{amsmath}
\usepackage{geometry}       
\usepackage{amsfonts}
\usepackage{framed}
\usepackage{float}

\newcolumntype{L}[1]{>{\raggedright\arraybackslash}m{#1}}
\newcolumntype{C}[1]{>{\centering\arraybackslash}m{#1}}
\newcolumntype{R}[1]{>{\raggedleft\arraybackslash}m{#1}}

\usepackage{color}
\makeatletter

\@addtoreset{equation}{section}
\makeatother 

\theoremstyle{plain}
\newtheorem{theorem}{Theorem}[section]

\theoremstyle{definition}

\newtheorem{remark}[theorem]{Remark}

\usepackage{latexsym}
 
\title[Topology of Currencies]{Topology of Currencies: Persistent Homology for FX Co-movements: A Comparative Clustering Study}

\author{Pattravadee de Favereau de Jeneret} \address{Maastricht University, School of Business and Economics, P.O.Box 616, 6200 MD, Maastricht, The Netherlands.} \email{m.defavereaudejeneret@student.maastrichtuniversity.nl}

\author{Ioannis Diamantis}
\address{Department of Data Analytics and Digitalisation, Maastricht University, School of Business and Economics, P.O.Box 616, 6200 MD, Maastricht, The Netherlands.} \email{i.diamantis@maastrichtuniversity.nl}

\keywords{Topological Data Analysis; Persistent Homology; Currency Co-movements; Clustering; Foreign Exchange (FX) Market; Clustering Methods; Wasserstein Distance}

\begin{document}

\begin{abstract}
This study investigates whether Topological Data Analysis (TDA) can provide additional insights beyond traditional statistical methods in clustering currency behaviours. We focus on the foreign exchange (FX) market, which is a complex system often exhibiting non-linear and high-dimensional dynamics that classical techniques may not fully capture. We compare clustering results based on TDA-derived features versus classical statistical features using monthly logarithmic returns of 13 major currency exchange rates (all against the euro). Two widely-used clustering algorithms, \(k\)-means and Hierarchical clustering, are applied on both types of features, and cluster quality is evaluated via the Silhouette score and the Calinski-Harabasz index. Our findings show that TDA-based feature clustering produces more compact and well-separated clusters than clustering on traditional statistical features, particularly achieving substantially higher Calinski-Harabasz scores. However, all clustering approaches yield modest Silhouette scores, underscoring the inherent difficulty of grouping FX time series. The differing cluster compositions under TDA vs. classical features suggest that TDA captures structural patterns in currency co-movements that conventional methods might overlook. These results highlight TDA as a valuable complementary tool for analysing financial time series, with potential applications in risk management where understanding structural co-movements is crucial.

\bigbreak

\noindent \textbf{JEL Classification:} C38, C63, F31, G15 \\
\noindent \textbf{MSC 2020:} 91B84, 62H30, 68T09, 91G70, 55N31
\end{abstract}

\maketitle

\section{Introduction}

In April 2022, trading in the over-the-counter foreign exchange (FX) market averaged more than \$7.5 trillion per day, representing an increase of 14\% compared to three years earlier \cite{ECB2023}. Given this large scale and the FX market's central role in global finance, even small fluctuations can have far-reaching effects across the world. For example, the 2008 global financial crisis highlighted how shocks in one currency can quickly spread to others. Understanding currency co-movements is therefore crucial for risk management, portfolio diversification, and mitigating systemic instability. Yet, currencies rarely move in isolation: their behaviour reflects a complex combination of macroeconomic, political, and structural factors.  

Reference rates play a central role in this system, serving as benchmarks for other financial instruments such as loans and exchange rates. They provide a common mechanism that links international interest payments and valuations to market conditions \cite{ECB2023}. Consequently, reference rates offer valuable information about currency co-movements, which can enhance systematic risk assessment.

Traditional statistical methods, such as correlation, covariance, and clustering, have been widely applied to uncover relationships between currency movements. These approaches are efficient and interpretable, but they rely on assumptions of linearity, low dimensionality, and stationarity that may not hold in the dynamic, interconnected FX market. As a result, they often fail to capture non-linear or higher-order dependencies, potentially underestimating systemic risk and misguiding policy or investment decisions.  

Topological Data Analysis (TDA), a recent framework grounded in algebraic topology, provides a complementary approach. TDA characterises the ``shape'' of data by studying how geometric and topological features evolve across scales \cite{Shultz2023}. Unlike traditional statistical tools, TDA does not depend on coordinate systems or linear assumptions, and it is robust to noise, making it well suited for analysing high-dimensional, non-linear, and complex systems such as the FX market.

Building on this foundation, the present study investigates the extent to which TDA can add value to classical statistical methods in the analysis of foreign-exchange dynamics. Specifically, we perform a comparative clustering study based on classical statistical features and TDA-derived features, evaluating whether TDA uncovers additional structure that conventional techniques may overlook.

\smallbreak 
Hence, the central research question is:

\begin{quote}
\textit{How does clustering based on TDA-derived features compare in quality to clustering based on classical statistical features?}
\end{quote}

The remainder of this paper is organised as follows.  
Section \ref{sec:literature} reviews the relevant literature on classical statistical and topological approaches to currency co-movements.  
Section \ref{sec:data} describes the data selection and preprocessing.  
Section \ref{sec:eda} presents an exploratory analysis of the dataset.  
Section \ref{sec:methods} outlines both statistical and TDA-based clustering procedures.  
Section \ref{sec:discussion} discusses the empirical results, limitations, and directions for future research.  
Finally, Section \ref{sec:conclusion} concludes the paper.

\section{Literature Review}
\label{sec:literature}

\subsection{Statistical Methods in FX Studies}

Numerous methods have been used to analyse currency co-movements and interrelationships, including correlation, covariance, and clustering. These approaches remain popular in finance due to their mathematical efficiency and interpretability. They can also be combined to provide a more comprehensive picture, with correlation and covariance often serving as foundational measures.

Mai \textit{et al.} \cite{Mai2017} employed network-based correlation analysis to study currency co-movements in the FX market, finding that currencies often cluster by geographic proximity. Their paper thus exemplifies both the strengths of correlation analysis, such as the derivation of regional modules, and its weaknesses, notably its dependence on linear relationships.  
Similarly, Dro?d? \textit{et al.} \cite{Drozdz2007} used correlation matrices to explore collective behaviour and clustering among 60 world currencies. Their findings indicated that results were strongly dependent on the choice of base currency: when the euro or US dollar served as the base, the derived co-movements were more diverse and revealed subtler interdependencies.  
Together, these studies show that correlation analysis can uncover meaningful patterns but remains limited by its linear assumptions.

Other studies have instead relied on covariance as a foundation. Andersen \textit{et al.} \cite{Andersen2003} used realised covariance to measure how pairs of asset returns move together, providing a valuable basis for improved volatility modelling and forecasting, an essential task in financial applications.

Building on these classical measures, clustering has become an increasingly common tool for analysing time-series behaviour in financial markets. Paparrizos \textit{et al.} \cite{Paparrizos2024} offer a comprehensive review of clustering methods, highlighting the relative strengths of approaches such as hierarchical clustering and emphasising the importance of selecting appropriate distance metrics. Traditional distance measures, they argue, often fail when directly applied to complex time-series data.

In financial applications, the typical input for clustering analyses is logarithmic returns, which are preferred for their stationarity and additive properties. For instance, Verma \textit{et al.} \cite{Verma2018} used log-returns to create log-volatility features, which were then clustered to capture empirical market dynamics.  
Similarly, Mantegna \cite{Mantegna1999} applied hierarchical tree analysis to price time series, showing that clusters tended to group companies operating within the same industry or sub-industry sectors, revealing meaningful economic structure through statistical similarity.

\subsection{Topological Data Analysis}

In recent years, Topological Data Analysis (TDA) has emerged as a versatile framework for uncovering the underlying \emph{shape} of data, offering tools that extend beyond traditional statistical measures.  
Garcia \cite{Garcia2022} provides an accessible introduction to persistent homology, demonstrating its capacity to extract structure from complex, high-dimensional data in applications ranging from natural-language processing to handwritten-digit recognition.

Lum \textit{et al.} \cite{Lum2013} illustrated TDA's broad applicability by extracting topological features across diverse domains, including cancer research, political science, and sports analytics, highlighting TDA's unique ability to detect subgroups and hidden relationships.  
El-Yaagoubi \textit{et al.} \cite{ElYaagoubi2023} further advanced this perspective by applying persistent homology to dependence networks in multivariate time series, with specific applications to brain connectivity. Their results showed that TDA can identify hidden cyclic structures through persistence landscapes, a property that translates naturally to financial data, where multivariate dependencies are common.

Within economics, TDA adoption has been limited but promising. Schultz \cite{Shultz2023} positioned the method as a complementary tool to econometrics, particularly well suited to non-linear and high-dimensional datasets where traditional models struggle. Among the potential applications discussed were early-warning indicators of market instability, an area where topology can reveal structural signals obscured by noise.

Aguilar \textit{et al.} \cite{Aguilar2020} used TDA to detect structural changes in US stock-market data, showing that persistent homology can capture early signs of instability and critical transitions. They cautioned, however, that the method remains computationally intensive and requires careful parameterisation of persistence landscapes.  
Majumdar \textit{et al.} \cite{Majumdar2020} demonstrated TDA's potential for clustering by combining it with self-organising maps (SOMs) and random forests on stock-price series. Their hybrid approach successfully distinguished simulated from real financial processes, underscoring TDA's ability to capture structures beyond volatility or correlation alone.

Recent comparative studies have extended these insights. Hobbelhagen \textit{et al.} \cite{Hobbelhagen2024} compared TDA with Symbolic Aggregate Approximation (SAX) on European stock markets, finding that while SAX offered greater scalability and interpretability for broad market trends, TDA provided finer insight into local stock movements.  
Building on this, Bereta \textit{et al.} \cite{Bereta2025} added the extended SAX (eSAX) approach and applied all three methods to consumer-behaviour time series. Their findings confirmed the complementarity of symbolic and topological methods: symbolic approaches were computationally efficient and interpretable, whereas TDA avoided ``catch-all'' clusters despite noisy data.

\subsection{Research Gap and Contribution}

Both statistical methods, such as covariance, correlation, and clustering, and TDA have yielded valuable insights in financial research. TDA, in particular, has proven effective in extracting meaningful structure from complex and noisy datasets, identifying early-warning signals of structural shifts, and modelling volatility. Despite this progress, applications of TDA within finance remain largely concentrated on stock-market data, with studies on the FX market still scarce. Moreover, most existing comparisons, such as those in \cite{Hobbelhagen2024} and  \cite{Bereta2025}, focus on symbolic representations rather than direct contrasts between TDA and conventional clustering approaches like \(k\)-means or hierarchical clustering.

Accordingly, this study addresses these gaps by offering a direct comparison of statistical and TDA-based clustering, specifically, \(k\)-means and hierarchical methods, within the context of the FX market.  
The results aim to reveal whether TDA captures non-linear dependencies and structural co-movements that traditional statistical techniques may overlook.

\section{Data Preparation}
\label{sec:data}

\subsection{Data Selection}

The data used in this study were collected from the European Central Bank (ECB) database. Multiple datasets were retrieved, each representing a time series of reference rates, defined as ``an exchange rate that is not intended to be used in any market transactions'' \cite{ECB2023}, with the euro serving as the base currency across all series.

Using reference rates rather than traded exchange rates reduces the influence of central bank interventions and emphasises factors such as global trade, inflation, monetary policy, and market activity. The ECB derives these reference rates primarily from actual market transactions in liquid markets with high trading volumes and frequent trades. For less liquid markets, the rates rely on firm bid/ask quotes, USD cross-rate calculations, and discretionary adjustments \cite{ECB2023}.

The use of a uniform base currency ensures a consistent and accurate comparison of relationships and movements among the different currencies. For instance, if volatility rises in some pairs while remaining stable in others, it may indicate regional or geopolitical shocks; conversely, a simultaneous rise across all pairs suggests systemic disruption centred around the euro area.

The ECB provides public access to 30 reference rates globally, including many European currencies not part of the euro area.  
For this study, a diverse set of currencies was selected to include both advanced and emerging economies:

\smallbreak

\begin{center}
AUD, BRL, CHF, CNY, GBP, INR, JPY, KRW, RUB, THB, TRY, USD, ZAR, and EUR.
\end{center}

The aggregated and cleaned dataset spans from 13 January 2000 to 1 March 2022, comprising 14 columns (excluding dates) and 5 993 observations. Each column represents a currency's reference rate time series, with EUR remaining constant as the numeraire.  
A second dataset was also constructed excluding the Russian ruble (RUB), extending the coverage to 27 November 2024 to capture more recent global events. These two datasets were later used in the exploratory analysis to better understand temporal behaviour and to contextualise clustering results.

Several currencies were selected because of their membership in the BRICS bloc (Brazil, Russia, India, China, South Africa), providing a robust environment for studying reference-rate behaviour under diverse macroeconomic regimes.  
To complement these, advanced economies (AUD, CHF, GBP, JPY, USD) were included, as they often act as global financial anchors and dominate international reserves. Changes in their rates can propagate globally, though the precise transmission dynamics require granular analysis.  
Finally, emerging-market currencies such as THB and KRW, which have shown relative stability, and TRY, which has undergone prolonged inflationary episodes since the late 2010s, were included to enrich the contrast between stability and volatility.

\subsection{Data Manipulation}

Before conducting the statistical and topological analyses, several preprocessing steps were necessary to ensure the comparability and stability of the currency time series. 
Foreign exchange reference rates exhibit heteroskedasticity, nonstationarity, and large variations in scale across currencies. 
To address these issues and enable meaningful clustering, the raw price levels were transformed into logarithmic returns, and the resulting series were standardised. 
These manipulations reduce the influence of differing volatility levels and align all currencies onto a common scale, ensuring that subsequent clustering and topological features reflect structural co-movements rather than differences in magnitude or scale.

\subsubsection{Logarithmic Returns}

Raw price levels, whether scaled or not, tend to cluster according to long-term trends rather than behavioural similarity \cite{Mattera2025}.  
To capture relative dynamics more accurately, the main analyses (Section \ref{sec:methods}) use monthly logarithmic returns.  
The euro was excluded since its constant value as the numeraire adds no information on relative movements and could generate a trivial, economically meaningless cluster.  
Moreover, in econometrics, returns typically satisfy the stationarity assumptions required by most time-series models better than price levels.  
The resulting dataset contains 13 columns (currencies) and 266 monthly observations, with varying numbers of daily data points per month aggregated accordingly.

\subsubsection{Standardisation}

Data scaling is a critical preprocessing step for both classification and distance-based clustering.  
De Amorim \textit{et al.} \cite{DeAmorim2022} and Wongoutong \cite{Wongoutong2024} showed that scaling can substantially affect model performance, with standardisation (Z-score normalisation) often outperforming min-max normalisation.  

Standardisation centres each feature at zero and rescales it so that its standard deviation equals one, ensuring equal weighting across currencies. This transformation preserves the shape of the data because it is linear, maintaining both order and relative distances.  
Unlike normalisation, it is unbounded and therefore more robust to outliers, an essential property for volatile financial time series.

This procedure was applied to the monthly log-returns dataset.  
Following Berthold {\it et al.} \cite{Berthold2016}, standardisation ensures that clustering is based on co-movement patterns rather than differences in magnitude or volatility.  
In this setting, Euclidean distances between standardised series approximate correlation distances, resulting in clusters that group currencies with similar directional movements rather than comparable variance alone.

\section{Exploratory Data Analysis}
\label{sec:eda}

\subsection{Time Series Analysis}

\begin{figure}[ht]
    \centering
    \begin{minipage}[b]{0.48\textwidth}
        \includegraphics[width=\linewidth]{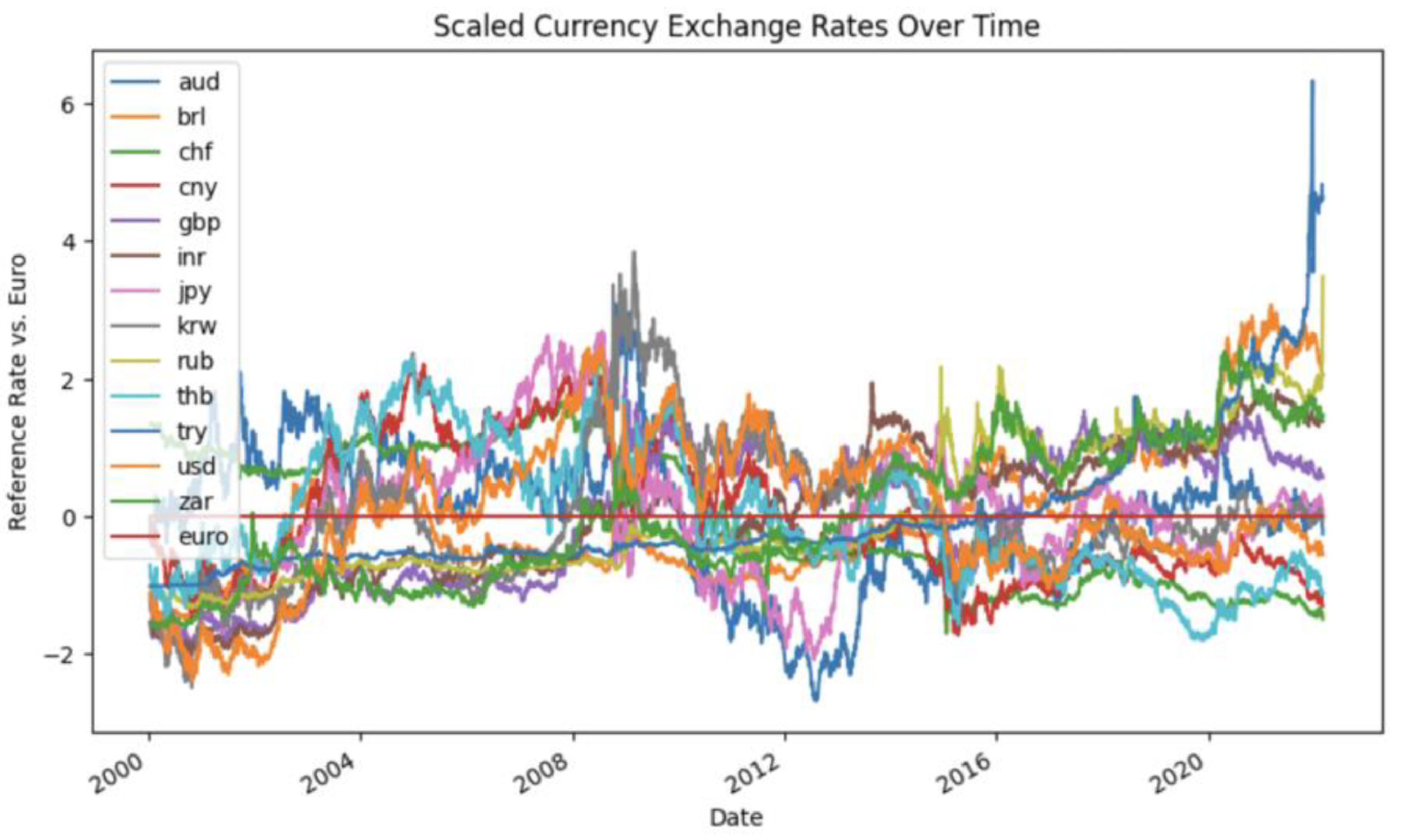}
    \end{minipage}
    \hfill
    \begin{minipage}[b]{0.48\textwidth}
        \includegraphics[width=\linewidth]{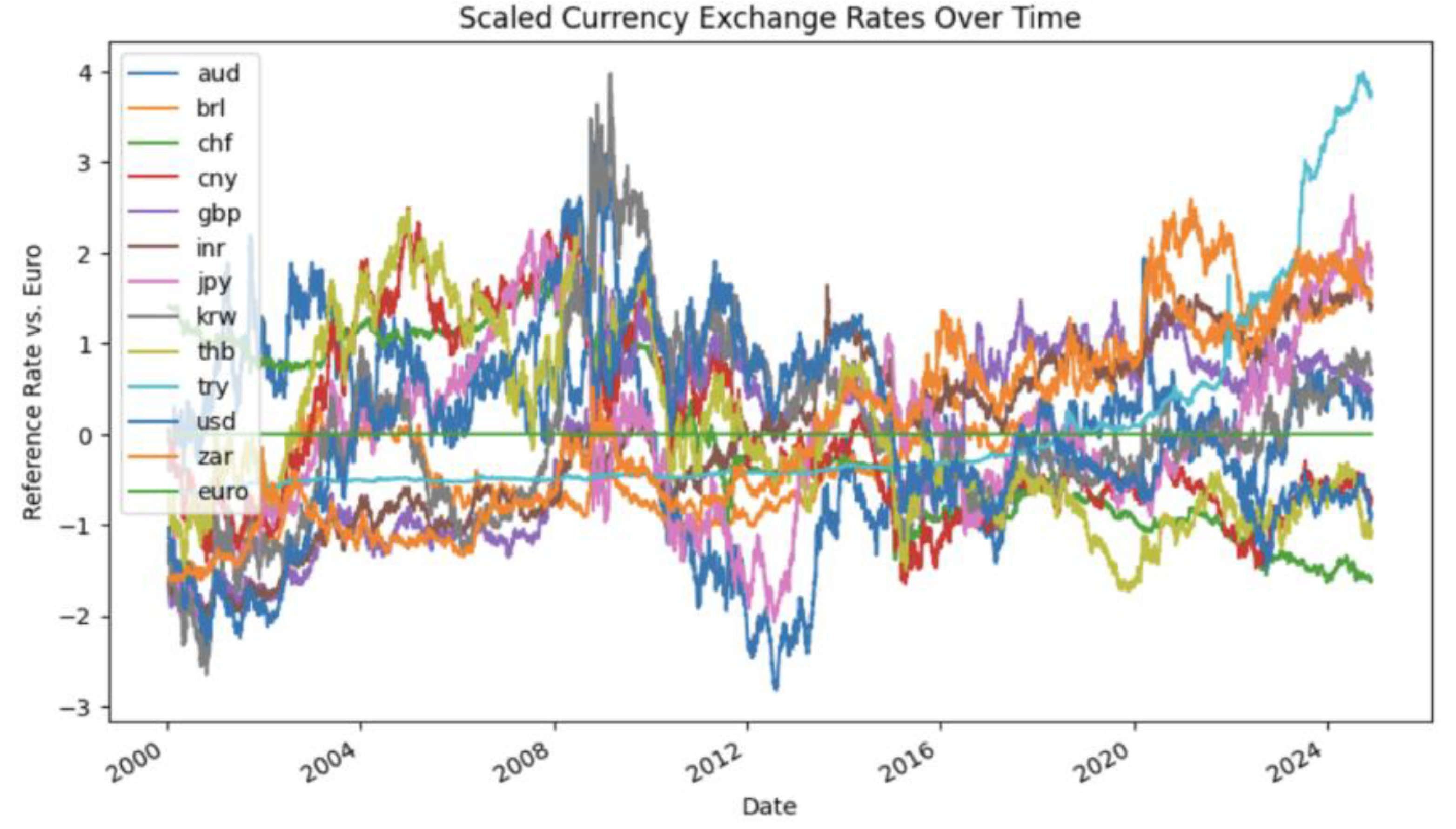}
    \end{minipage}
    \caption{Comparison of standardised currency movements across two time periods. On the left 2000--2022, and on the right 2000--2024.}
    \label{fig:scaled}
\end{figure}

Figure~\ref{fig:scaled} presents the standardised movements of all currencies over the respective sample periods.  
These plots enable the visual identification of economic and geopolitical patterns.  
A general depreciation of the euro is visible between 2009 and 2012, when most currencies trend downward simultaneously.  
In contrast, when different currencies deviate in direction, this suggests that localised factors dominate rather than systemic eurozone effects.

An interesting feature is that the vertical scale (standardised z-scores) contracts slightly in the longer dataset ending in~2024.  
This is most likely driven by the extreme inflation of the Turkish lira (TRY) combined with the scaling procedure:  
since standardisation rescales each feature to unit variance, the inclusion of more high-valued data points increases the pre-scaled standard deviation, reducing the resulting z-scores and thereby compressing the apparent spread.

\subsection{Variance}

\begin{figure}[ht]
    \centering
   \begin{minipage}[b]{0.26\textwidth}
        \includegraphics[width=\linewidth]{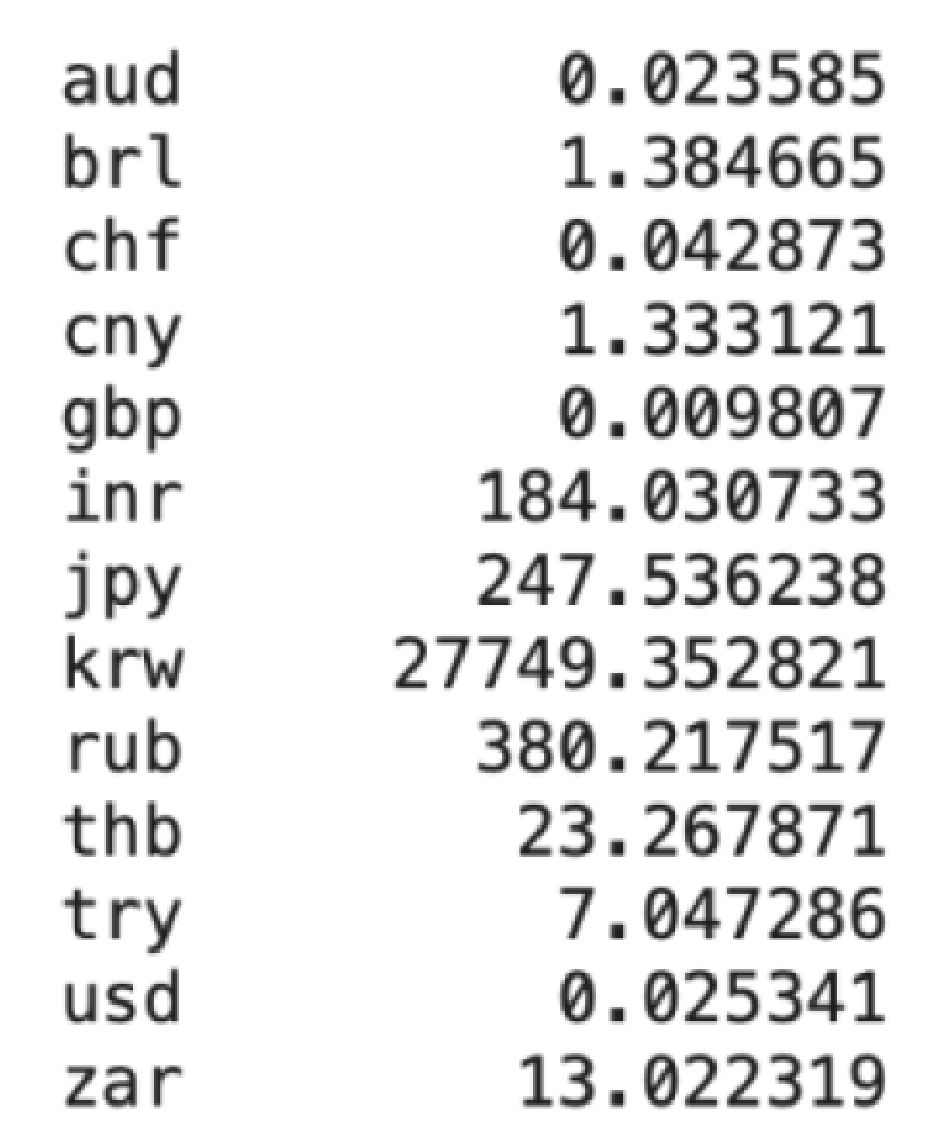}
    \end{minipage}
\qquad
   \begin{minipage}[b]{0.28\textwidth}
        \includegraphics[width=\linewidth]{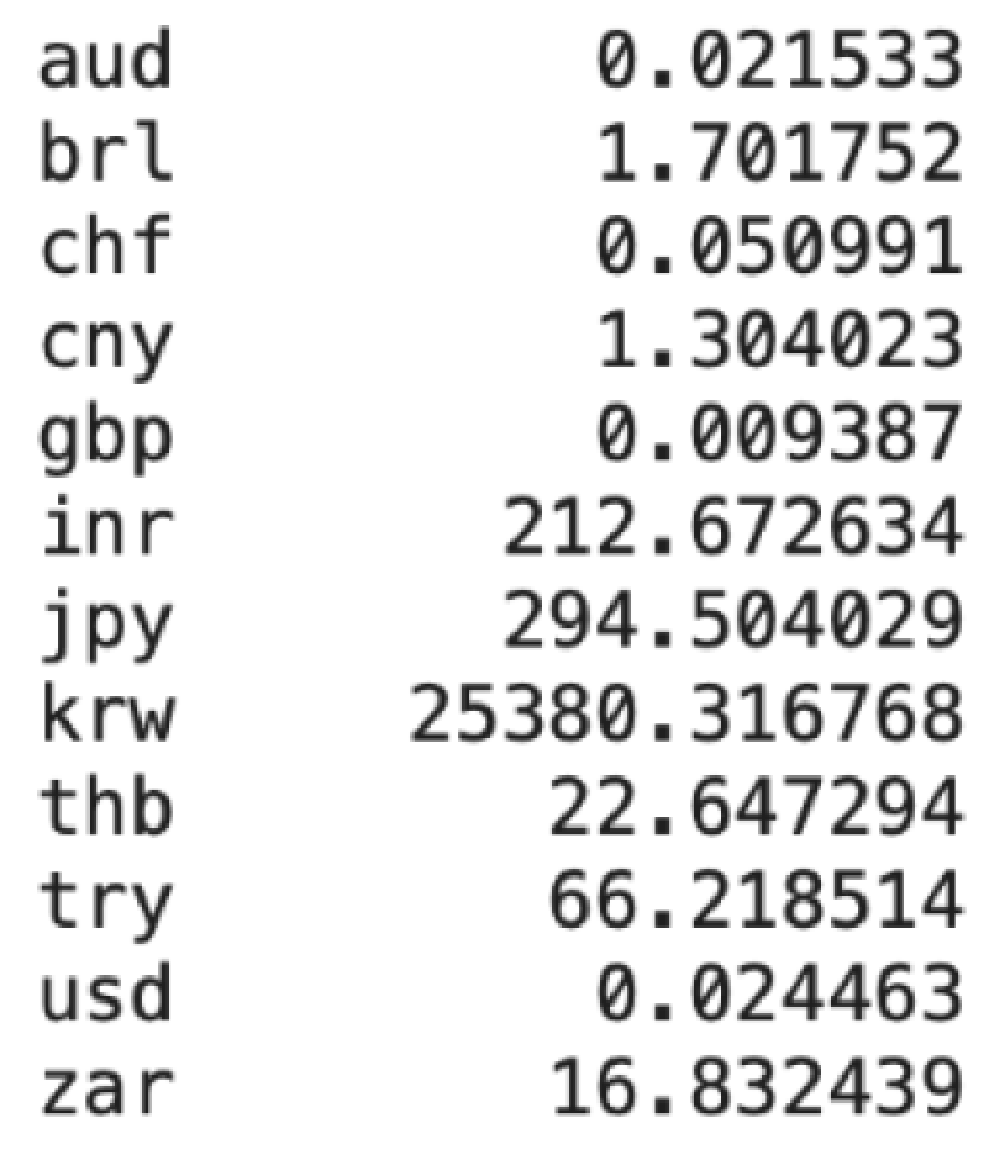}
    \end{minipage}
    \caption{Comparison of variance of currencies across the same two periods. On the left 2000--2022, and on the right 2000--2024.}
    \label{fig:variance}
\end{figure}

To complement the time-series plots, Figure~\ref{fig:variance} displays the variance of each currency's log-returns for both sample periods.  
The earlier dataset (2000--2022) and the extended dataset (2000--2024) show that most currencies exhibit only modest shifts in variance.  
Currencies such as the AUD, CNY, GBP, KRW, THB, and USD experienced slight decreases in volatility, whereas the BRL, CHF, INR, JPY, TRY, and ZAR displayed higher variance.

The TRY stands out, showing a more than ninefold increase in variance over just two and a half years, signalling a prolonged phase of extreme volatility and macroeconomic instability.  
While variance captures overall dispersion, it does not reveal time-dependent relationships; hence, further temporal decomposition was performed.

\subsection{Time Series Decomposition}

Although variances summarise dispersion, they ignore temporal ordering.  
To investigate structural patterns over time, each series was decomposed into trend, seasonal, and residual components using the Seasonal-Trend decomposition via Loess (STL) method \cite{Theodosiou2011}.  
This approach extends classical additive and multiplicative models, which assume fixed seasonal patterns and smooth trends, by employing locally weighted regression (LOESS) to fit evolving local structures.  
STL decomposition therefore offers robustness to outliers and adaptability to non-stationary economic data.

\begin{figure}[ht]
    \centering
    \includegraphics[width=0.7\textwidth]{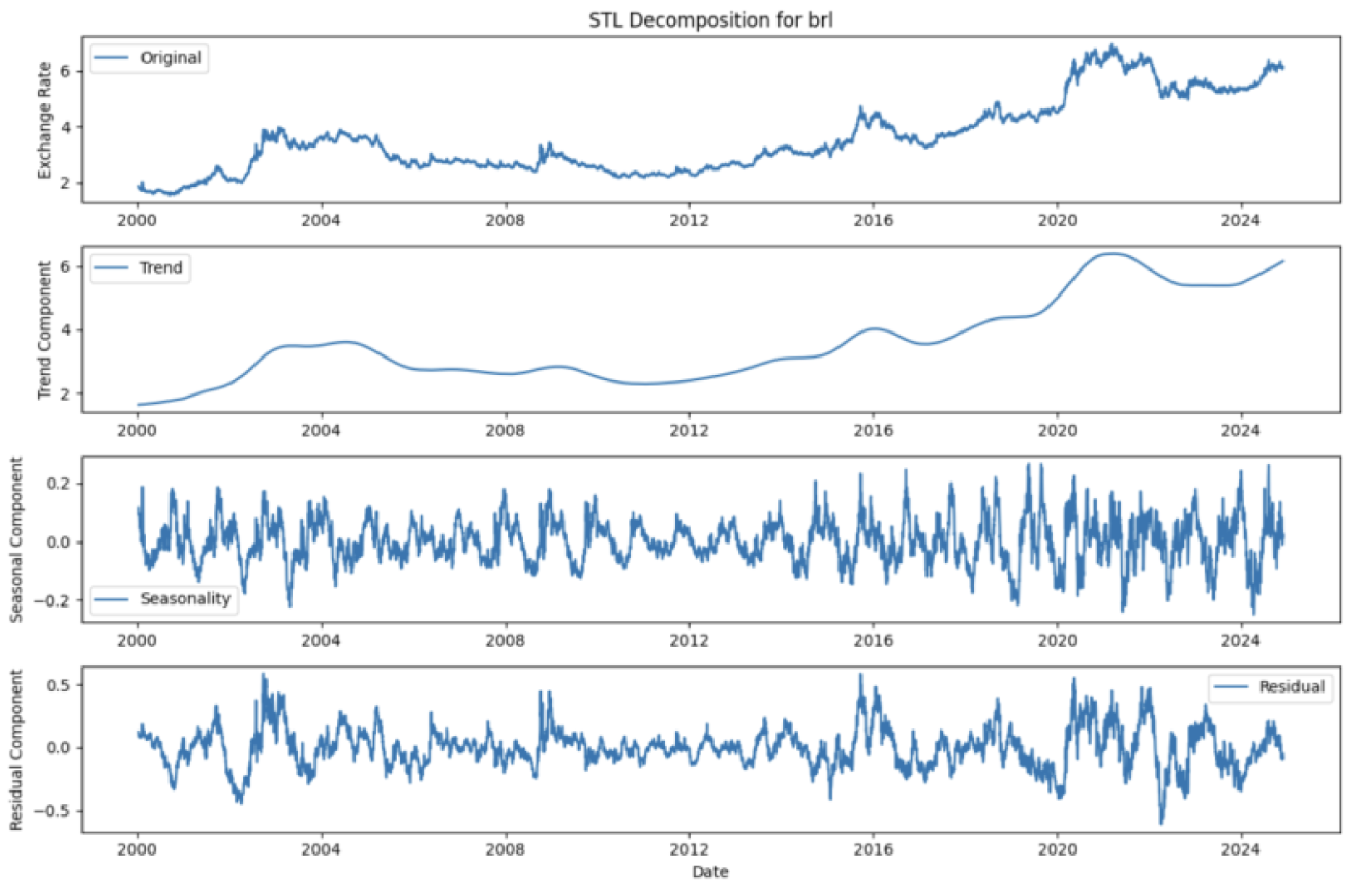}
    \caption{STL decomposition of the Brazilian real (BRL).}
    \label{fig:stl_brl}
\end{figure}

Figure~\ref{fig:stl_brl} shows the decomposition of the Brazilian real (BRL).  
The long-term trend reveals cyclical appreciation and depreciation phases, punctuated by large structural shifts associated with political and global economic events, such as the 2008 financial crisis and the COVID-19 pandemic.  
Residual spikes are observed around 2002, 2008, 2016, 2020, and 2022, aligning with major domestic disruptions including the corruption scandal and presidential impeachment of 2016 \cite{Watts2017} and the 2022 election-related uncertainty \cite{ISPI2022}.

\begin{figure}[ht]
    \centering
    \includegraphics[width=0.7\textwidth]{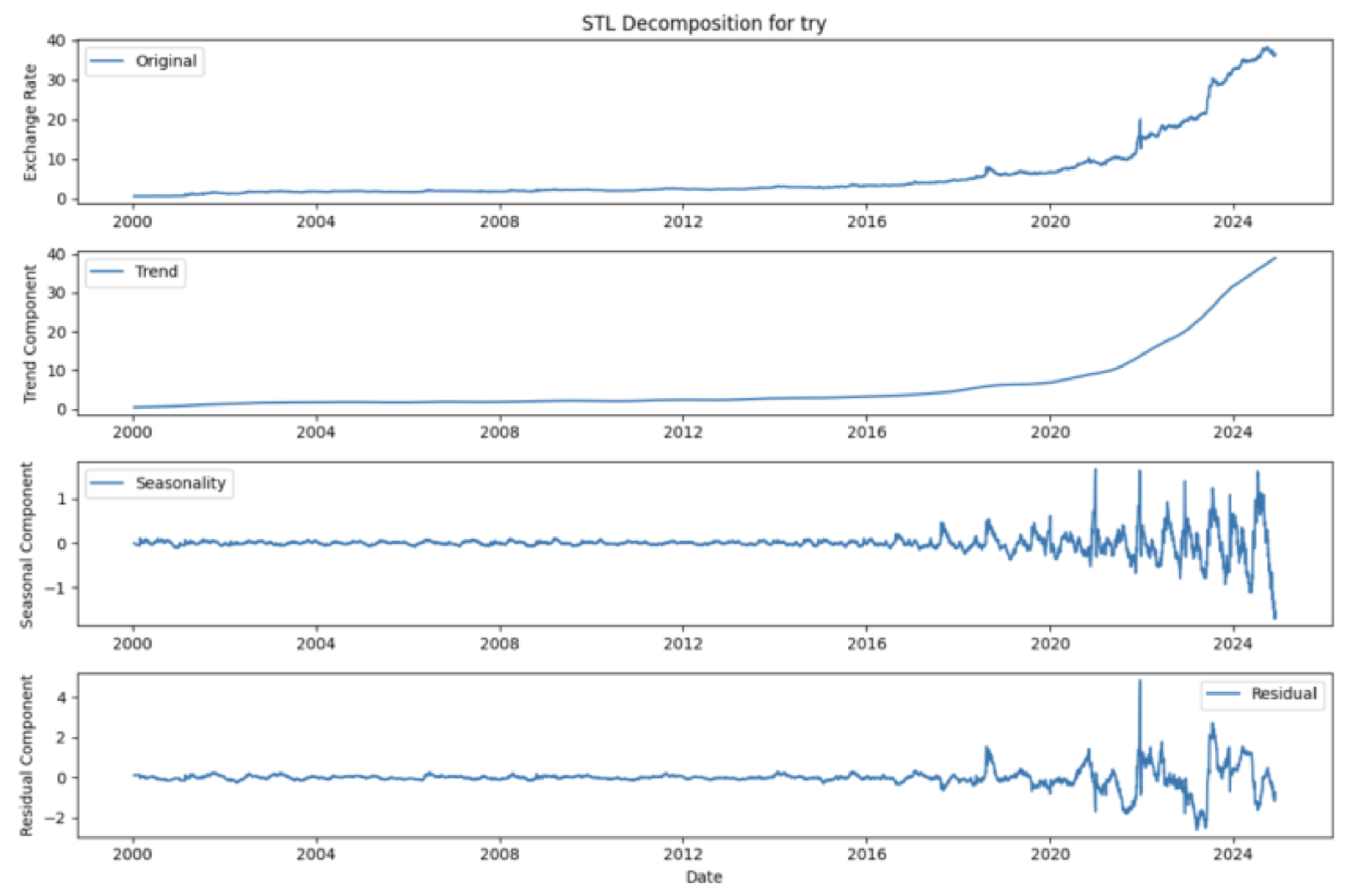}
    \caption{STL decomposition of the Turkish lira (TRY).}
    \label{fig:stl_try}
\end{figure}

Figure~\ref{fig:stl_try} presents the Turkish lira (TRY), whose decomposition displays an abrupt structural shift around 2018 following an extended period of relative stability.  
This sharp change corresponds to unorthodox monetary policies, including rate cuts that accelerated inflation and undermined investor confidence \cite{Pierini2021}.  
The seasonal component exhibits disrupted cycles, further emphasising the economy's structural transformation.

\begin{figure}[ht]
    \centering
    \includegraphics[width=0.7\textwidth]{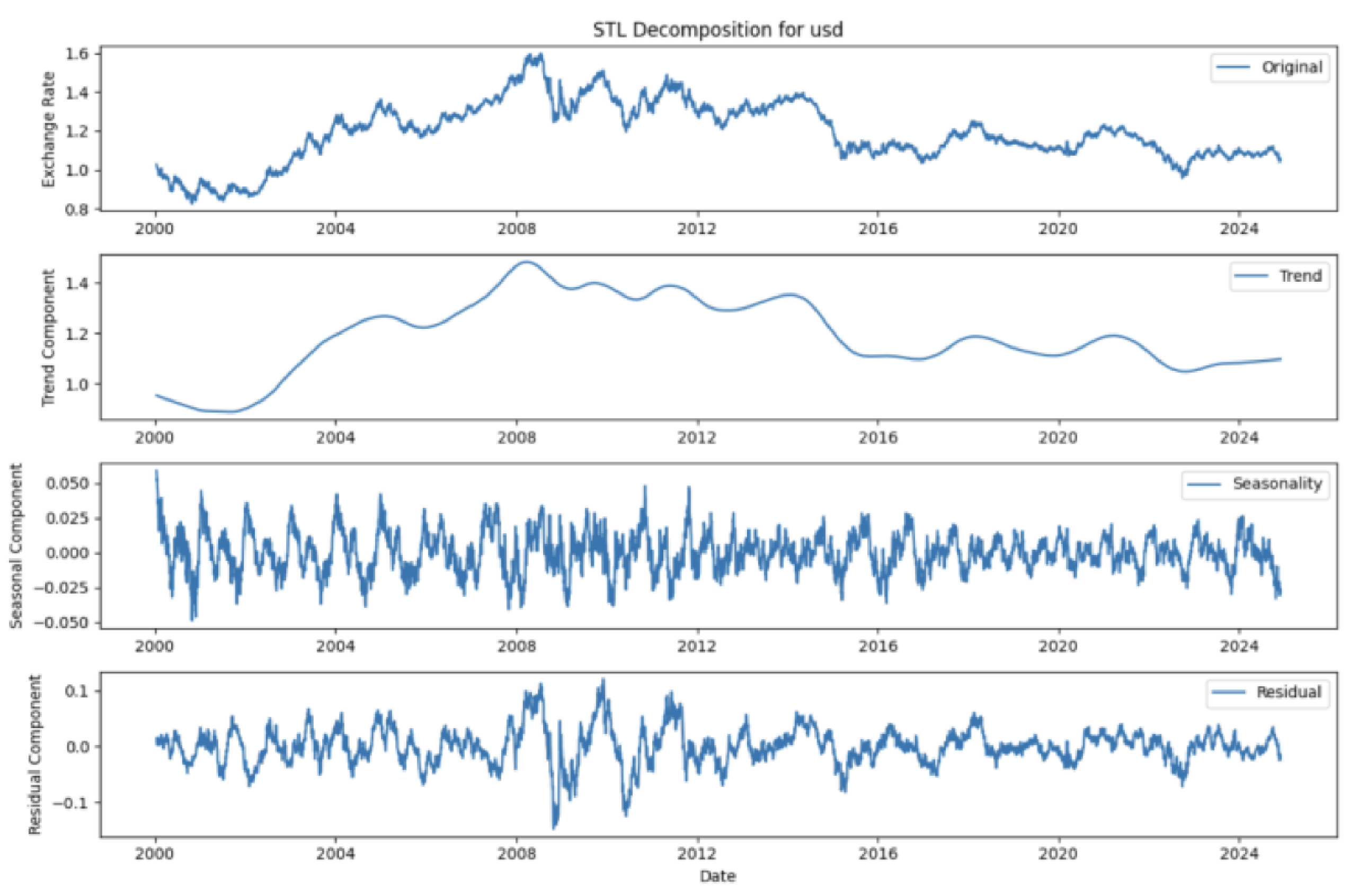}
    \caption{STL decomposition of the US dollar (USD).}
    \label{fig:stl_usd}
\end{figure}

Finally, Figure~\ref{fig:stl_usd} displays the decomposition for the US dollar (USD), which acts as a baseline of relative stability.  
Its trend shows moderate fluctuations, consistent with the dollar's role as a global safe-haven currency.  
Unlike emerging-market counterparts, its residual component is small and relatively stable through time.

\section{Methods and Analysis}
\label{sec:methods}

This section presents the analytical framework adopted to examine currency co-movements through both classical statistical and topological approaches. 
The analysis proceeds in two main stages. 
First, traditional econometric tools are applied to the standardised time series to establish a statistical benchmark based on covariance, correlation, and clustering. 
Second, Topological Data Analysis (TDA) is introduced to extract structural and nonlinear features from the same data using persistent homology and Wasserstein distances. 
By comparing the clustering outcomes obtained from these two fundamentally different representations, the study aims to evaluate whether TDA can reveal latent relationships in the foreign exchange market that are not captured by linear statistical measures.

\subsection{Classical Analysis}

The classical component of the analysis establishes a benchmark based on well-known statistical measures of dependence and similarity. 
It focuses on covariance and correlation structures among currency returns and applies standard clustering algorithms, namely, \(k\)-means and hierarchical agglomerative clustering, to these statistical features. 
This provides a linear, interpretable baseline against which the added value of topological methods can later be assessed.

\subsubsection{Covariance Matrix}

Covariance quantifies joint variability in both direction and magnitude, providing a first view of co-movements that can inform clustering. We compute a $13\times 13$ covariance matrix on monthly log-returns. As expected, self-pairs equal one. The largest off-diagonal covariance is for CNY-USD ($\approx 0.96$), implying closely aligned movements; this is consistent with China's management of the yuan relative to the dollar \cite{Anstey2024}. From a euro-based perspective, EUR-USD and EUR-CNY offer limited diversification if their fluctuations are highly similar.

CNY also covaries strongly with THB (0.75) and INR (0.72), potentially reflecting regional trade linkages; USD shows a similar pattern. The highest \emph{average} covariance across pairs is for THB (0.474), followed by INR (0.471), suggesting broad alignment with other series (more global forces, fewer idiosyncratic shocks). At the other end, CHF-TRY is the only negative pair ($-0.08$), pointing to diversification potential. CHF generally exhibits low covariances (e.g., with RUB 0.02; AUD 0.05; BRL 0.05; ZAR 0.09), consistent with an independent policy stance and safe-haven role from a European vantage.

\begin{figure}[ht]
  \centering
  \includegraphics[width=0.65\textwidth]{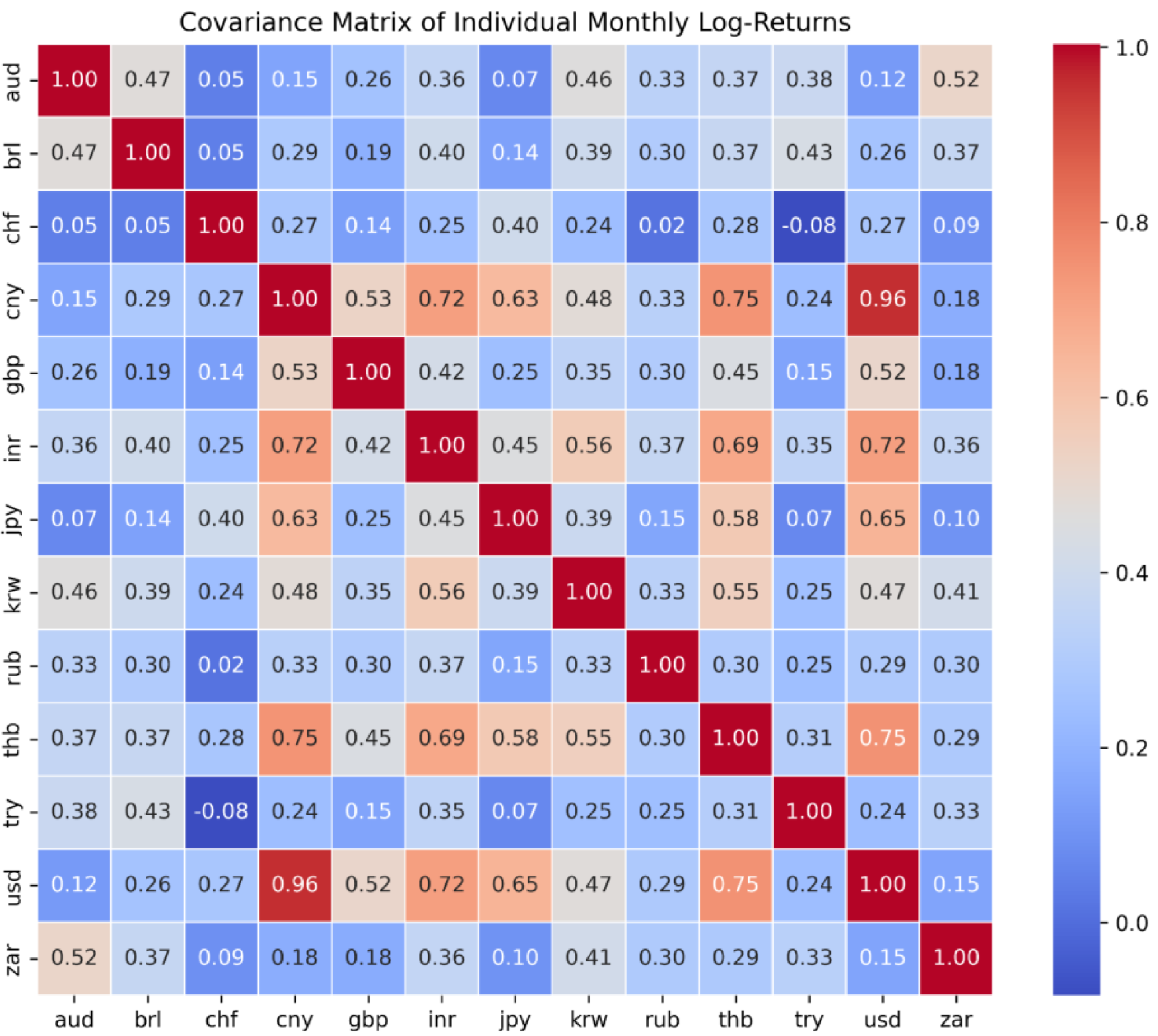}
  \caption{Covariance heatmap of monthly log-returns (13 currencies).}
  \label{fig:cov_heatmap}
\end{figure}

Covariance alone, however, conflates magnitude with direction and ignores temporal lead-lag. While standardising variables makes the covariance numerically comparable to Pearson correlation, alternative correlation measures can reveal distinct aspects of dependence.

\begin{remark}[On covariance versus correlation]\rm
Because all time series were standardised to zero mean and unit variance before computing the covariance matrix,
the resulting values are numerically equivalent to Pearson correlation coefficients. 
Throughout this paper, the term ``covariance'' is used in a broad sense to emphasize joint variation, 
but readers should note that, under standardisation, the two measures coincide up to a constant scaling factor. 
This does not affect any inference or clustering results but is important for conceptual precision.
\end{remark}

\subsubsection{Correlation Matrices}

Pearson correlation assumes linearity, continuity, approximate normality, and contemporaneous relationships \cite{Schober2018}. We therefore also examine Spearman correlation (monotone but potentially nonlinear associations \cite{DeWinter2016}) and \emph{cross-correlation} with lags up to one month, which can capture delayed responses \cite{Conlon2008}. The cross-correlation matrix is most appropriate for monthly time series in this setting.

Most pairs behave similarly across covariance/correlation (see for example CNY-USD and CHF-TRY), but notable exceptions exist. AUD-JPY, for example, shows 0.07 covariance versus $-0.12$ cross-correlation, indicating that lagged responses can flip the apparent relationship: movements are not fully synchronised, and one series' shifts can be followed by the other moving in the opposite direction.

\begin{figure}[ht]
  \centering
  \includegraphics[width=0.65\textwidth]{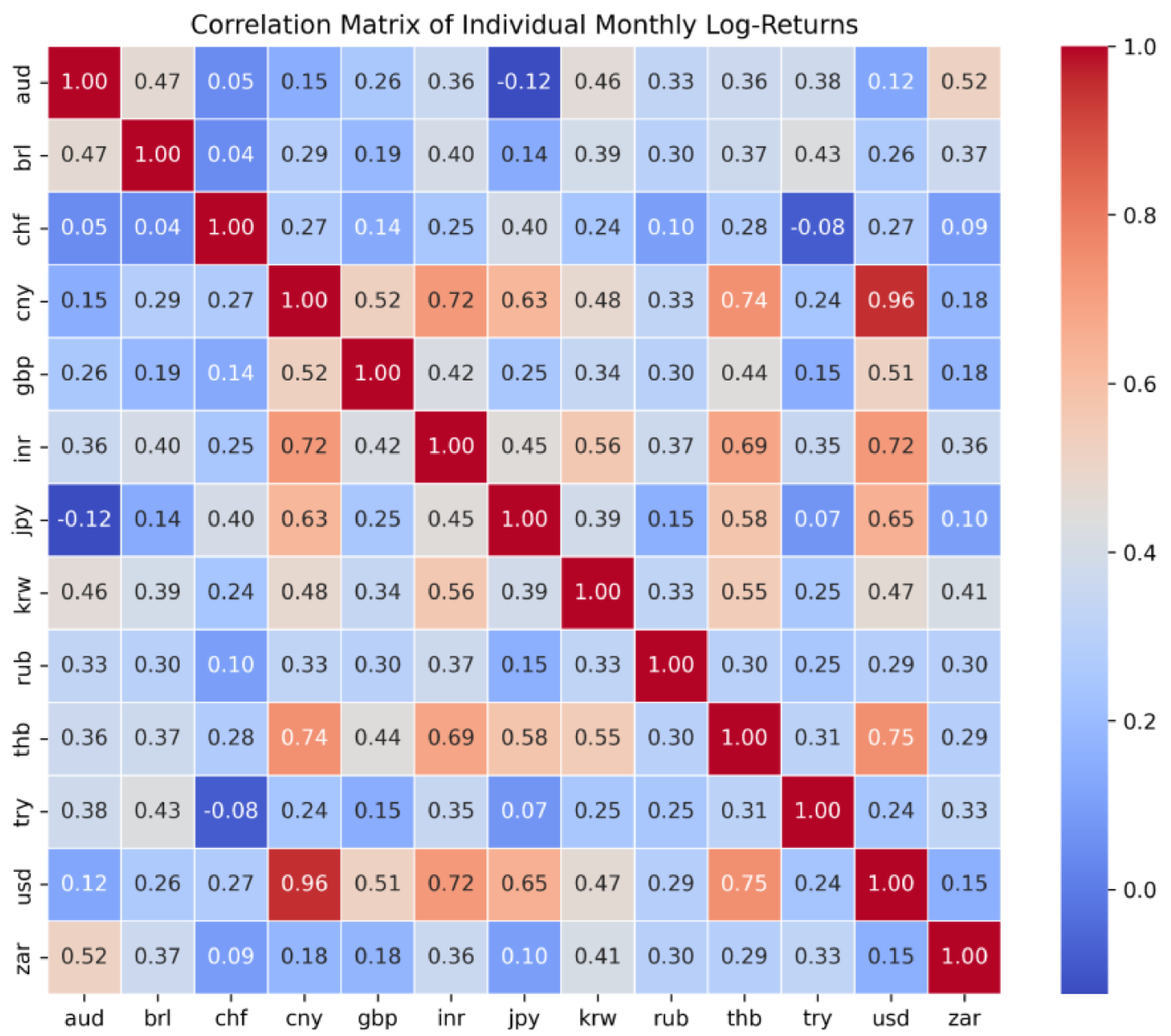}
  \caption{Cross-correlation heatmap of monthly log-returns (max lag 1 month).}
  \label{fig:xcorr_heatmap}
\end{figure}

\begin{remark}[On the interpretation of cross-correlations]\rm
The reported cross-correlation coefficients correspond to the maximum absolute correlation 
within a one-month lag window. 
While this approach captures short-term lead-lag effects, it can inflate dependence estimates 
when multiple testing across lags is unaccounted for. 
No formal significance testing was applied here, as the focus was exploratory; 
future analyses could assess lag significance via Bartlett's formula 
and adjust for multiple comparisons to obtain confidence bounds for the strongest correlations.
\end{remark}

\subsubsection{Clustering Setup and Metrics}

We cluster currencies to detect naturally occurring groups based on behavioural similarity. We use two unsupervised algorithms: \(k\)-means and agglomerative hierarchical clustering. Cluster quality is evaluated with:

\begin{itemize}
    \item[(i)] the average Silhouette coefficient (range \([-1,1]\); larger is better \cite{Rousseeuw1987}) and
    \item[(ii)] the Calinski-Harabasz (CH) index, which balances between-cluster separation and within-cluster cohesion \cite{Wang2021}. CH is comparative rather than absolute.
\end{itemize}

\noindent {\bf \(k\)-means on statistical features.}
Each currency is represented by its standardised monthly return vector. We select \(k=3\) via the elbow method (as in \cite{Bereta2025}).

\begin{figure}[ht]
  \centering
  \includegraphics[width=0.65\textwidth]{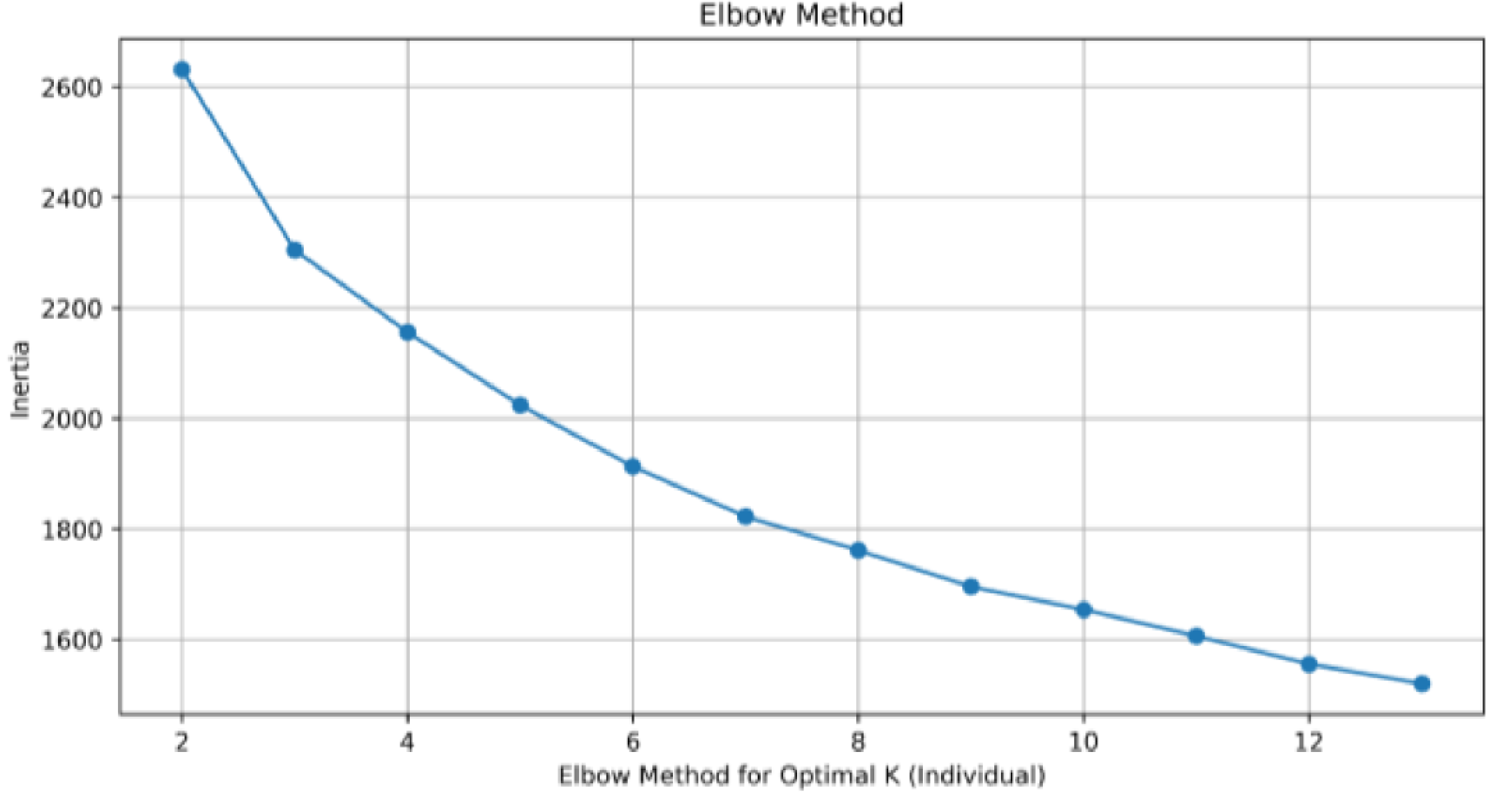}
  \caption{Elbow method for selecting \(k\) on statistical features.}
  \label{fig:elbow_stat}
\end{figure}

\begin{table}
\centering
\caption{Statistical \(k\)-means clusters (standardised return histories).}
\label{tab:stat_kmeans}
\begin{tabular}{ll}
\toprule
\textbf{Cluster} & \textbf{Currencies} \\
\midrule
1 & GBP \\
2 & CHF, CNY, INR, JPY, KRW, THB, USD \\
3 & AUD, BRL, RUB, TRY, ZAR \\
\bottomrule
\end{tabular}
\end{table}

Cluster~1 is a GBP singleton, consistent with idiosyncratic shocks (e.g., Brexit) and structural breaks relative to the euro. Cluster~2 mixes advanced (CHF, JPY, USD) and relatively stable emerging currencies (CNY, INR, KRW, THB), plausibly reflecting steadier distributions and, for some, managed regimes. Cluster~3 (AUD, BRL, RUB, TRY, ZAR) exhibits lower correlations to others and trend components showing persistent depreciation/appreciation, often tied to domestic instability and commodity exposure.  
Scores: Silhouette $0.110$; CH $2.657$.

\smallbreak 
\noindent {\bf Hierarchical clustering on statistical features.}
We use complete linkage with Euclidean distance. The resulting dendrogram differs from \(k\)-means: GBP is no longer a singleton; RUB appears boundary-like (late merge). INR/KRW/THB remain near CNY/ USD, consistent with regional linkages.

\begin{figure}[ht]
  \centering
  \includegraphics[width=0.7\textwidth]{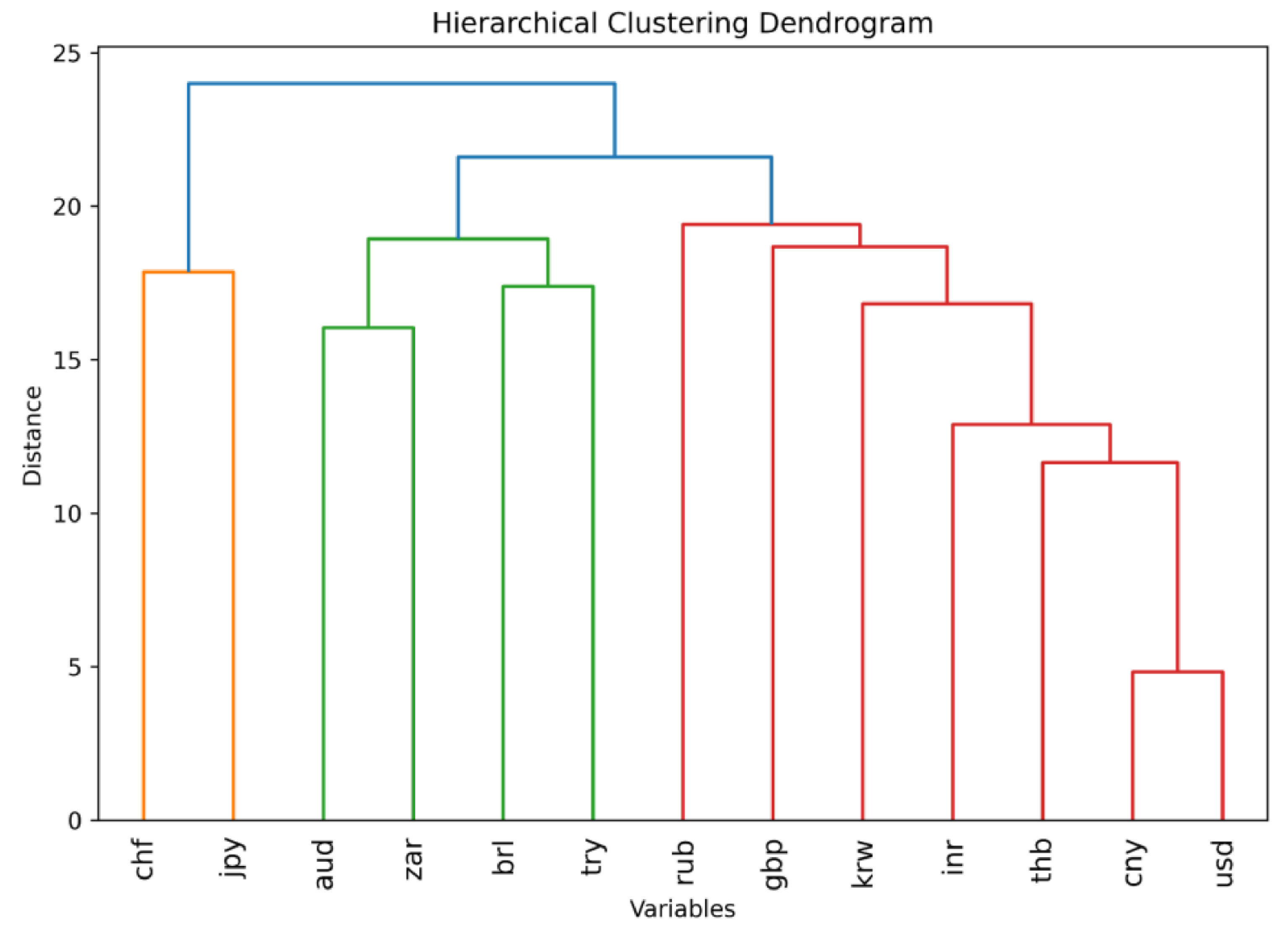}
  \caption{Hierarchical clustering.}
  \label{fig:stat_dendro}
\end{figure}

CHF and JPY form a tight pair despite their cross-correlation sensitivity, aligning with their shared safe-haven role.  
Scores: Silhouette $0.111$; CH $2.942$.

\subsection{Topological Data Analysis (TDA)}

TDA characterises the \emph{shape} of data via persistent homology, tracking when topological features (components $H_0$, loops $H_1$, higher-order cavities) appear and disappear across a filtration scale. In our context, shapes can signal cycles, structural breaks, or regime switches not captured by linear contemporaneous measures.

\subsubsection{Sliding-Window Embedding and Point Clouds}

\begin{figure}[ht]
  \centering
  \includegraphics[width=0.6\textwidth]{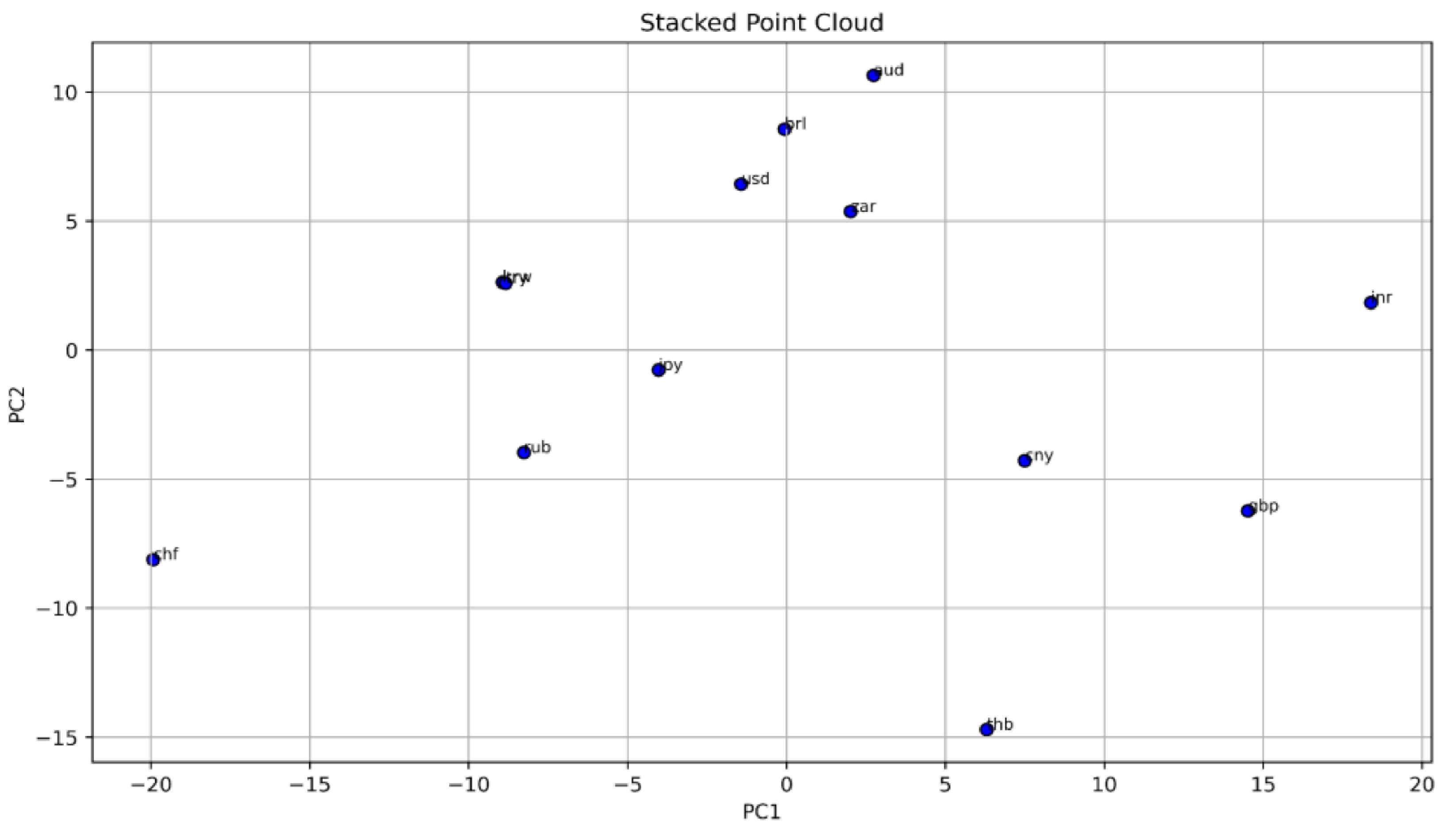}
  \caption{Two-dimensional visualisation (PCA) of stacked topological summaries across currencies.}
  \label{fig:pca_pointclouds}
\end{figure}

Each univariate time series is mapped into $\mathbb{R}^d$ using a sliding window with window length $d$ and delay $\tau$ \cite{Perea2013}. Smaller windows capture local variation; larger windows can capture full cycles but risk noise in high dimensions. Following diagnostics (false nearest neighbours/autocorrelation, cf.\ \cite{Truong2017}), most currencies favour $(d,\tau)=(4,1)$; we adopt these for consistency across series and the pairwise set. This yields 13 point clouds (one per currency). Topological summaries are stacked and projected for visualisation via PCA (Figure~\ref{fig:pca_pointclouds}).

\subsubsection{Mathematical Background: Persistent Homology for Embedded Time Series}
\label{sec:math_ph}

Let $x_1,\ldots,x_T \in \mathbb{R}$ denote a univariate time series (here: monthly log-returns for a currency). 
A sliding-window (delay) embedding with window length $d \in \mathbb{N}$ and delay $\tau \in \mathbb{N}$ produces a point cloud
\[
X \;=\; \Big\{\, \mathbf{v}_t \;=\; 
\big(x_t,\,x_{t-\tau},\,\dots,\,x_{t-(d-1)\tau}\big) \in \mathbb{R}^d 
\;\;:\;\; t=(d-1)\tau+1,\dots,T \,\Big\},
\]
equipped with a metric $d(\cdot,\cdot)$ (we use Euclidean). Intuitively, $X$ encodes local temporal patterns as geometric structure in $\mathbb{R}^d$.

\smallbreak 

\noindent {\bf Vietoris-Rips filtration.}
For $\varepsilon \ge 0$, the Vietoris-Rips complex $\mathrm{VR}_\varepsilon(X)$ is the abstract simplicial complex whose $k$-simplices are $(k\!+\!1)$-point subsets of $X$ with all pairwise distances $\le \varepsilon$; increasing $\varepsilon$ grows the complex:
\[
\mathrm{VR}_{\varepsilon} (X) \;\subseteq\; \mathrm{VR}_{\varepsilon'}(X)\quad \text{for }\varepsilon \le \varepsilon'.
\]
The nested family $\{\mathrm{VR}_\varepsilon(X)\}_{\varepsilon \ge 0}$ is a \emph{filtration} that tracks how connectivity and higher-order relations appear as we thicken points into simplices \cite{Edelsbrunner2010}.

\smallbreak

\noindent {\bf Homology and persistent homology.}
For each $\varepsilon$, the $k$-th homology group $H_k\big(\mathrm{VR}_\varepsilon(X)\big)$ (with coefficients in a field, e.g.\ $\mathbb{Z}_2$) is the quotient
\[
H_k\big(\mathrm{VR}_\varepsilon(X)\big) \;=\; \ker \partial_k \big/ \operatorname{im} \partial_{k+1},
\]
where $\partial_k$ are the simplicial boundary maps of the chain complex associated with $\mathrm{VR}_\varepsilon(X)$.
Elements of $H_0$ are connected components; elements of $H_1$ are loops (1-dimensional cycles); higher $H_k$ capture higher-dimensional voids.

Filtration inclusions induce linear maps on homology,
\[
\iota_{\varepsilon \le \varepsilon'}:\;
H_k\big(\mathrm{VR}_\varepsilon(X)\big)\;\longrightarrow\; H_k\big(\mathrm{VR}_{\varepsilon'}(X)\big),
\]
so $\{H_k(\mathrm{VR}_\varepsilon(X)), \iota_{\varepsilon \le \varepsilon'}\}$ forms a \emph{persistence module}. 
Under mild finiteness assumptions (finite metric space $X$), this decomposes into interval modules; equivalently, each topological feature has a \emph{birth} scale $b$ and \emph{death} scale $d$ \cite{Edelsbrunner2010}. 
This yields:
\begin{itemize}
  \item the \emph{barcode}: a multiset of intervals $[b_i,d_i)$, and
  \item the \emph{persistence diagram} $D_k(X)$: a multiset of points $(b_i,d_i)$ in the birth--death plane.
\end{itemize}
Features farther from the diagonal ($d-b$ larger) are more \emph{persistent} and typically regarded as more structurally meaningful.

\smallbreak 

\noindent {\bf Distances and stability.}
To compare shapes, we use metrics between diagrams, notably the $p$-Wasserstein distance $W_p$ (or the bottleneck distance as $p\!\to\!\infty$), which computes the optimal matching cost between birth--death pairs (with the diagonal as a sink for unmatched mass). 
Crucially, persistence diagrams are \emph{stable} with respect to perturbations of the underlying distances: small changes in $X$ (e.g.\ noise, minor resampling) lead to small changes in $D_k(X)$ under $W_p$ \cite{Edelsbrunner2010}. 
This robustness is especially desirable when analysing financial time series, which are inherently noisy and non-stationary.

\smallbreak 

\noindent {\bf Why this matters for FX dynamics.}
The delay-embedded cloud $X$ turns \emph{temporal behaviour} into \emph{geometry}. 
Persistent homology then interprets that geometry as topological structure:
\begin{itemize}
  \item $H_0$ (components): long-lived components indicate \emph{persistent segmentation} of local patterns (e.g.\ regime shifts or structural breaks where return patches stay distinct over a wide scale $\varepsilon$). 
        Rapid $H_0$ merging suggests homogeneous local dynamics.
  \item $H_1$ (loops): persistent loops encode \emph{recurrent} or \emph{cyclic} behaviour in the embedded patterns (e.g.\ seasonality, oscillatory mean-reversion, business-cycle effects). 
        Short-lived loops signal weak or noisy cycles; long-lived loops indicate strong, stable cyclical structure.
\end{itemize}
In summary, persistent homology captures aspects of currency behaviour that classical contemporaneous correlation cannot: (i) \emph{shape} and recurrence of local trajectories, (ii) \emph{scale} at which structures appear/disappear, and (iii) \emph{robustness} of those structures to noise. 
In our application, pairwise Wasserstein distances between diagrams quantify how similar two currencies are in their \emph{structural dynamics}, not merely in linear co-movements. 
Those distances are then used for clustering (directly for hierarchical clustering; via Euclidean embedding for \(k\)-means), providing a complementary partition to standard statistical features.

This mathematical formulation provides the foundation for the computational procedure described next, where we construct the persistence modules explicitly from the embedded currency trajectories using Vietoris-Rips filtrations.

\subsubsection{Complexes and Filtrations}

We build Vietoris-Rips complexes on pairwise distances, growing simplices as the filtration threshold $\varepsilon$ increases. 

Alternative complexes (such as alpha or Delaunay–Rips) can more faithfully capture the geometric structure of the underlying point cloud because they rely on exact proximity relations rather than pairwise thresholds. However, these constructions become computationally demanding and numerically less stable in higher dimensions, where simplicial growth scales combinatorially \cite{Edelsbrunner2010,Mishra2023}.

\begin{figure}[ht]
  \centering
  \begin{minipage}[b]{0.47\textwidth}
    \includegraphics[width=\linewidth]{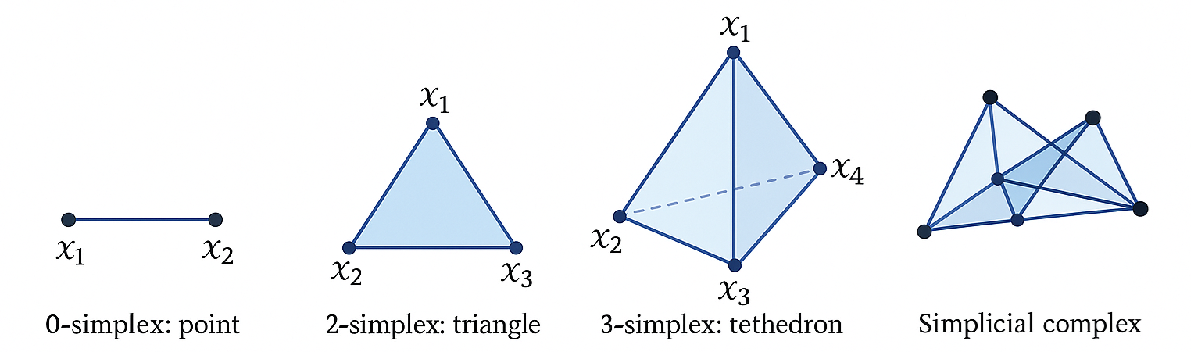}
  \end{minipage}\hfill
  \begin{minipage}[b]{0.47\textwidth}
    \includegraphics[width=\linewidth]{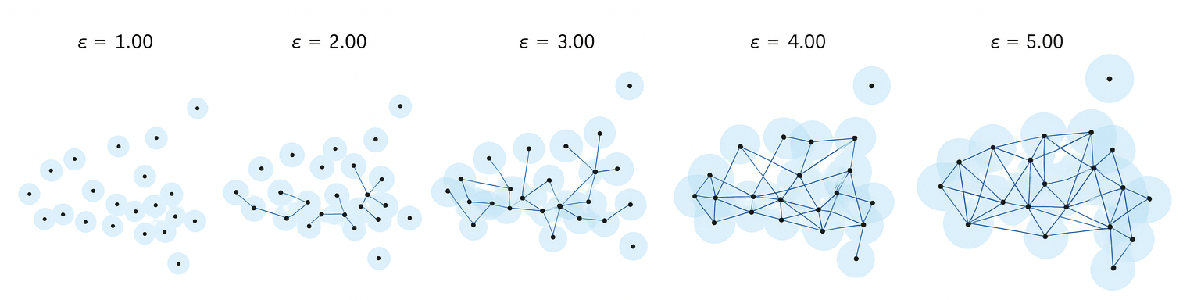}
  \end{minipage}
  \caption{Topological building blocks and filtration process.}
  \label{fig:simplices_filtration}
\end{figure}

\subsubsection{Persistence Diagrams and Barcodes}\label{subsec:perdiagr}

Persistence diagrams plot \((\text{birth},\text{death})\) of features; points far from the diagonal are more persistent (thus more structurally meaningful). Barcodes provide an equivalent interval view.

\begin{figure}[ht]
  \centering
  \begin{minipage}[b]{0.32\textwidth}
    \includegraphics[width=\linewidth]{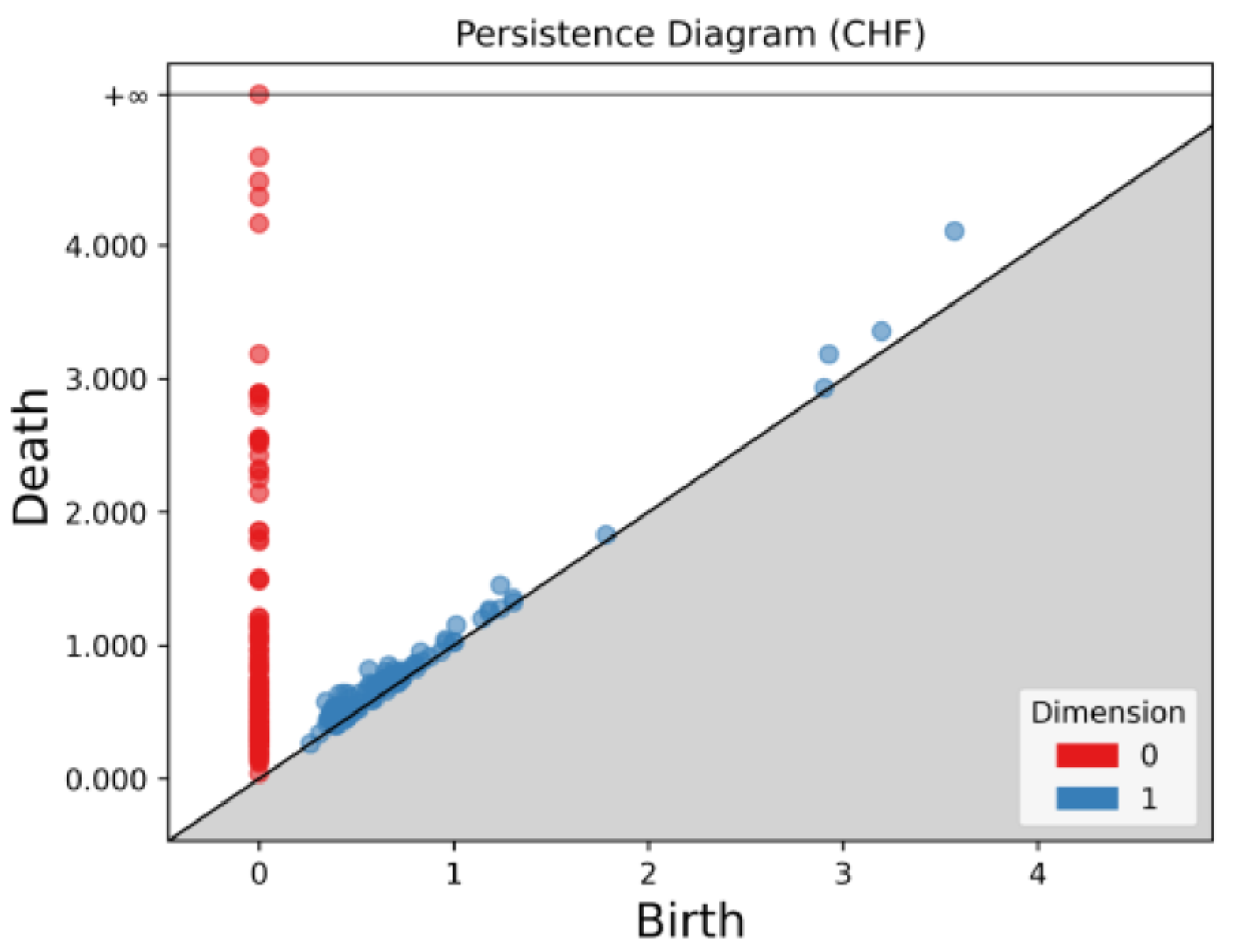}
  \end{minipage}
  \begin{minipage}[b]{0.32\textwidth}
    \includegraphics[width=\linewidth]{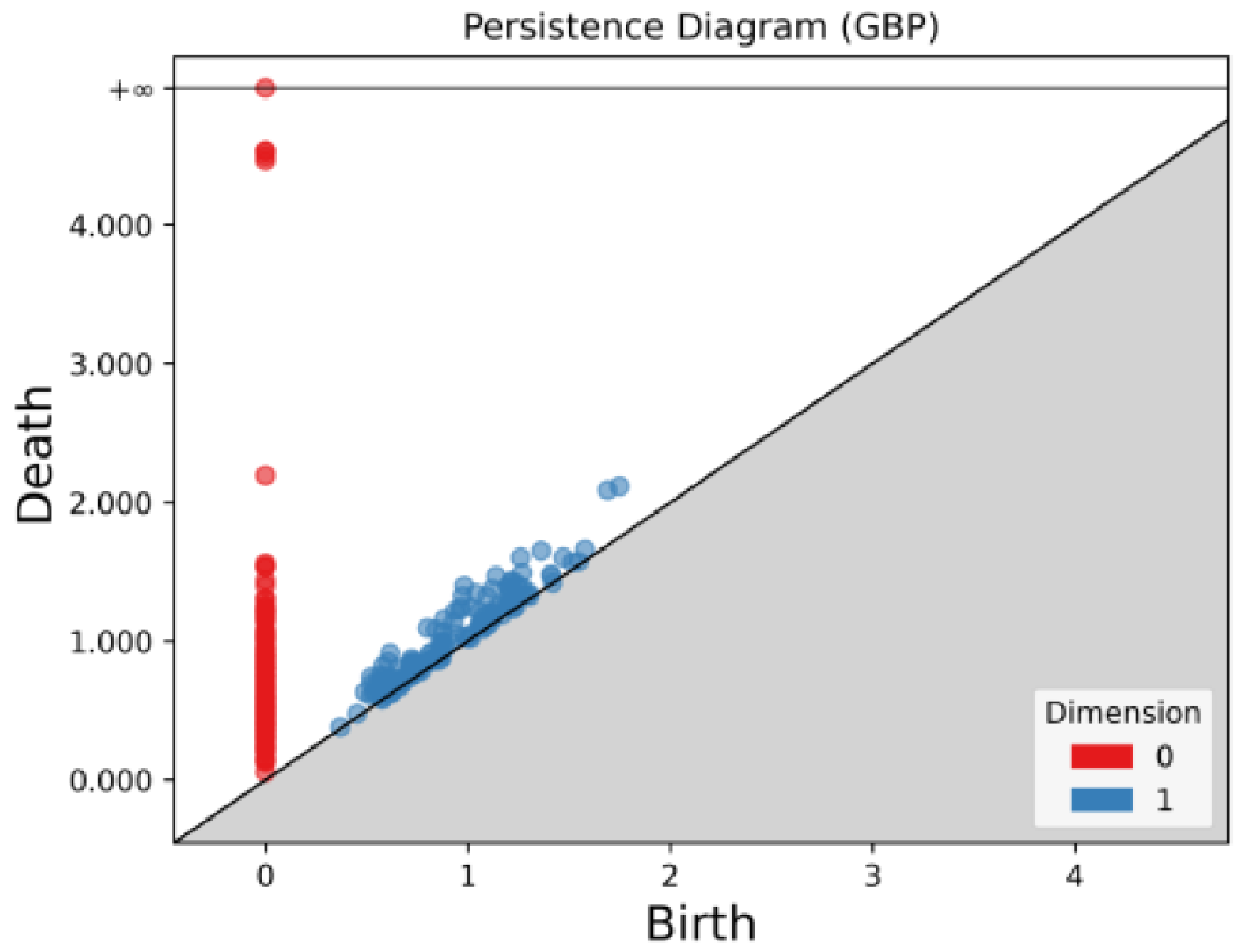}
  \end{minipage}
  \begin{minipage}[b]{0.32\textwidth}
    \includegraphics[width=\linewidth]{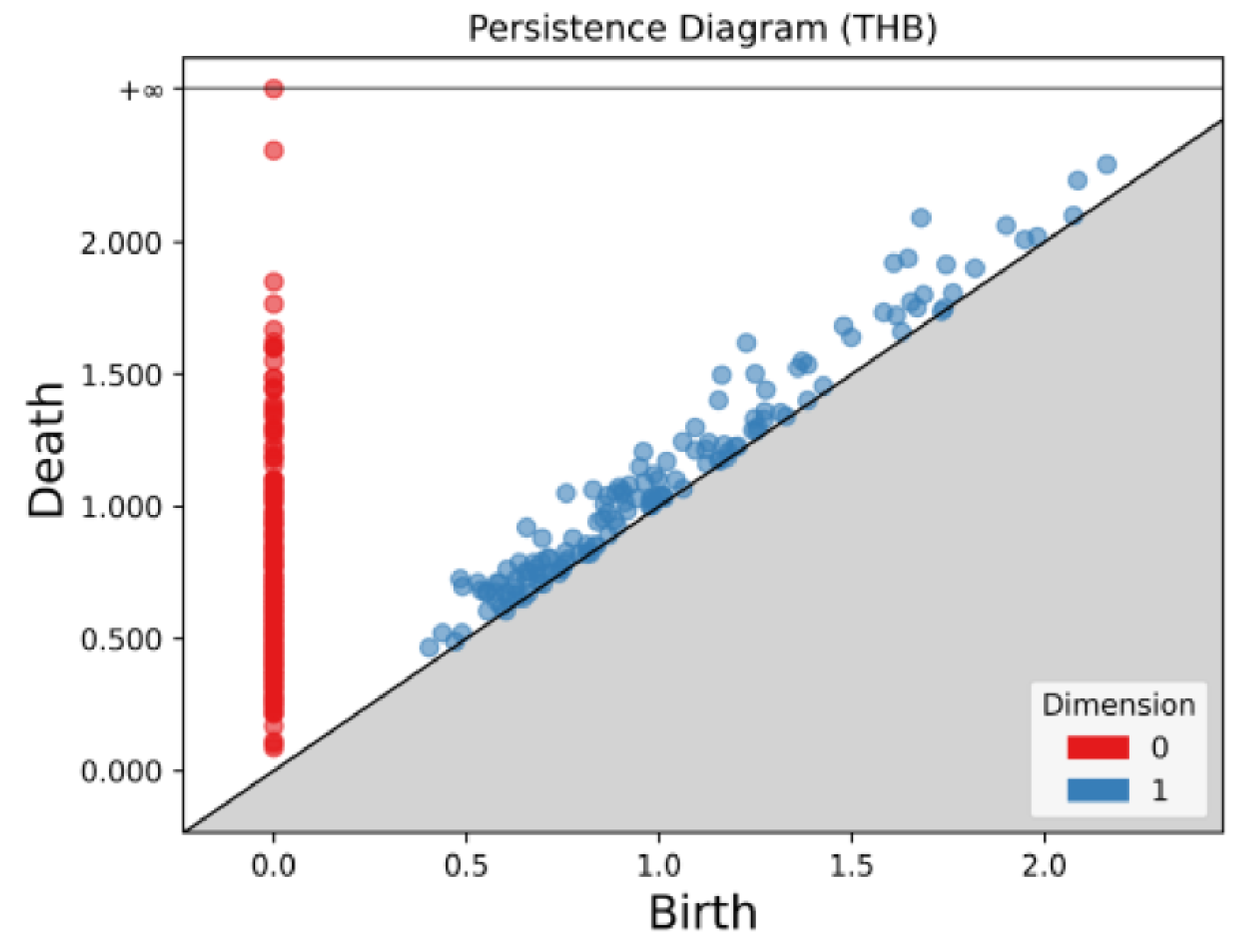}
    \label{fig:pd_thb}
  \end{minipage}
  \caption{Persistence diagrams ($H_0$ red, $H_1$ blue) for CHF, GBP and.THB}
  \label{fig:pd_examples}
\end{figure}

GBP's $H_0$ deaths cluster near the diagonal (rapid merging), with a notable fragmentation episode (structural breaks/outliers). CHF shows an intermediate $H_0$ profile and multiple persistent $H_1$ loops (strong cyclicity). THB's features are short-lived and close to the diagonal, consistent with homogeneous/managed dynamics.

\begin{figure}[ht]
  \centering
  \begin{minipage}[b]{0.32\textwidth}
    \includegraphics[width=\linewidth]{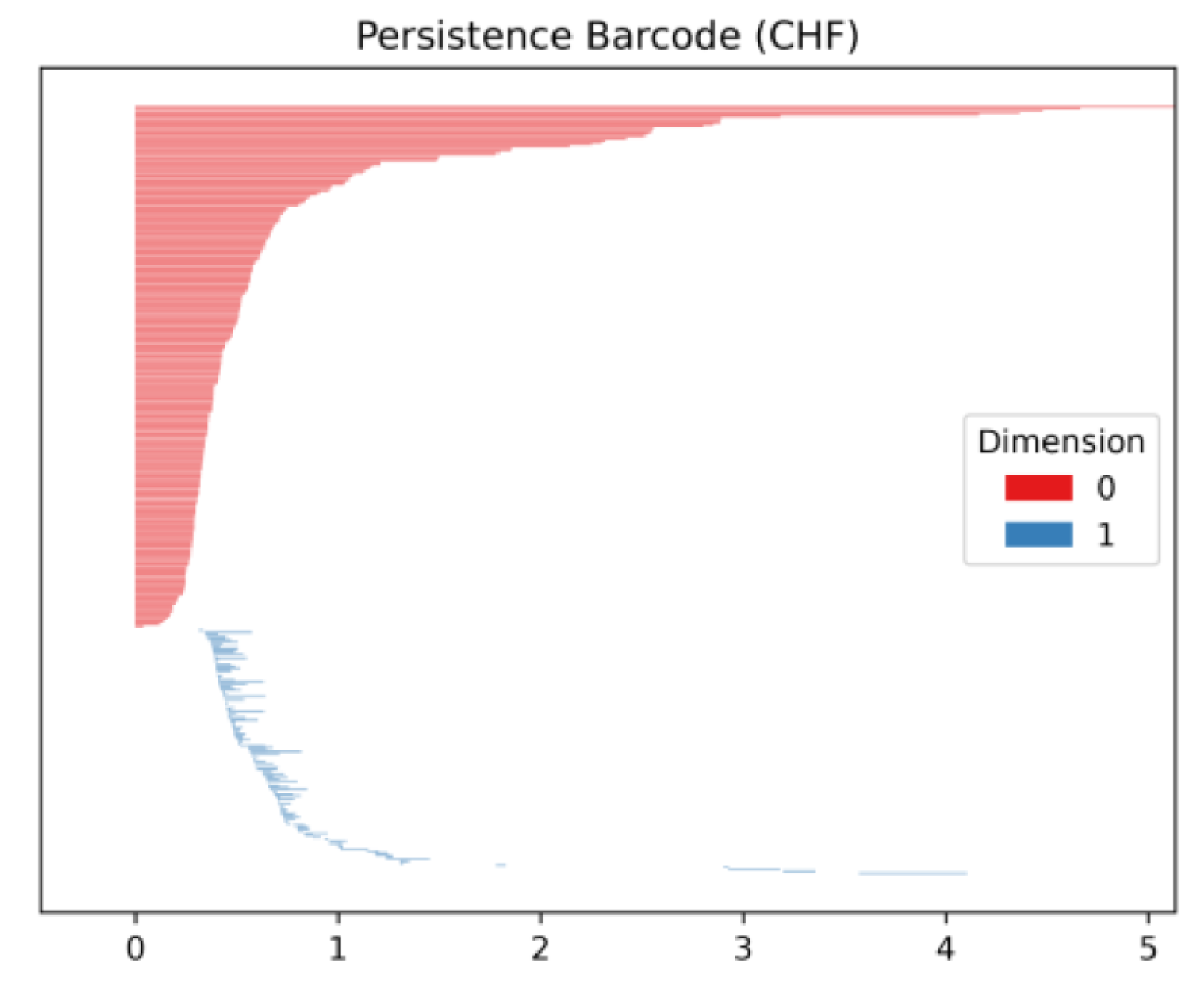}
  \end{minipage}
  \begin{minipage}[b]{0.32\textwidth}
    \includegraphics[width=\linewidth]{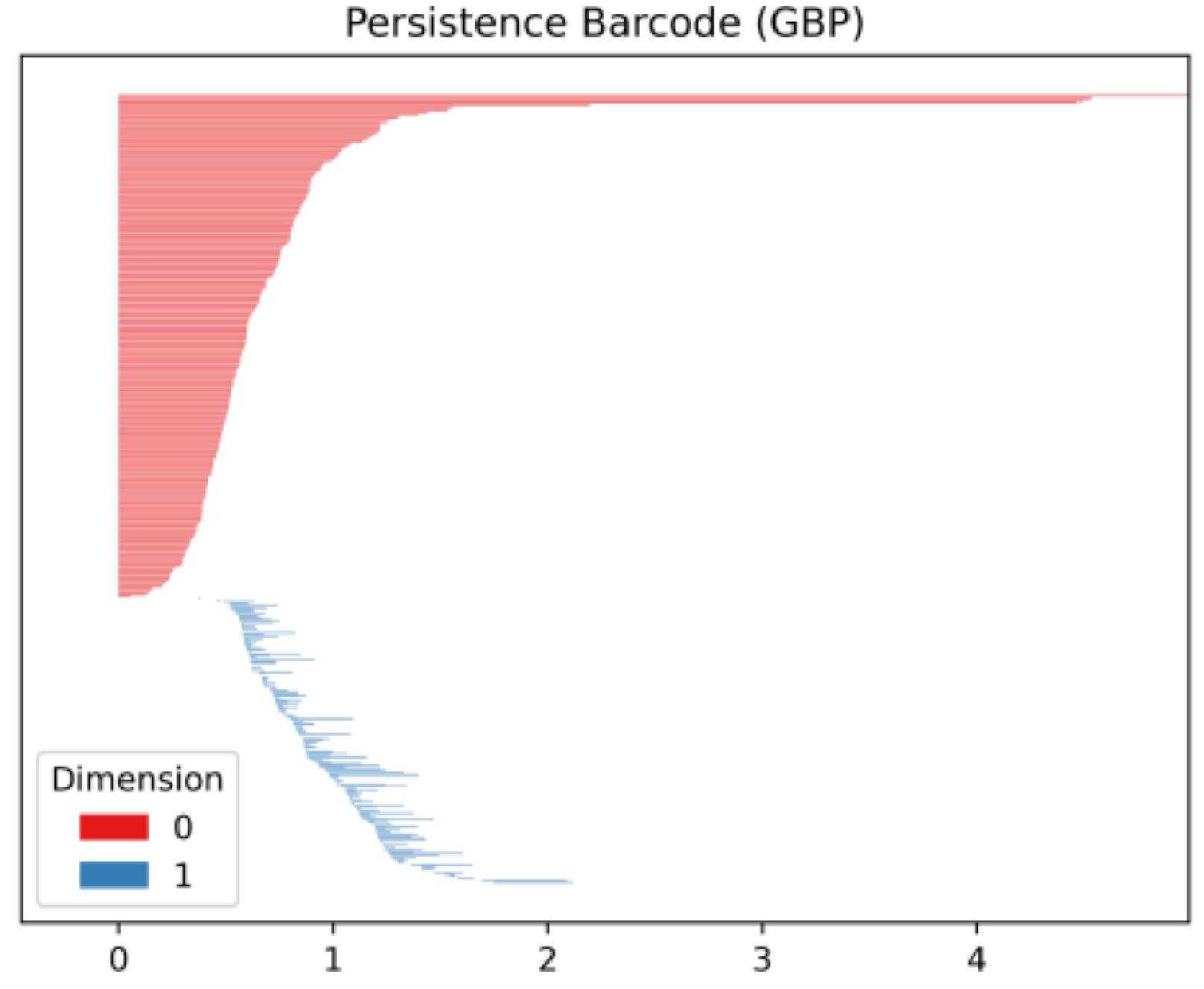}
  \end{minipage}
  \begin{minipage}[b]{0.32\textwidth}
    \includegraphics[width=\linewidth]{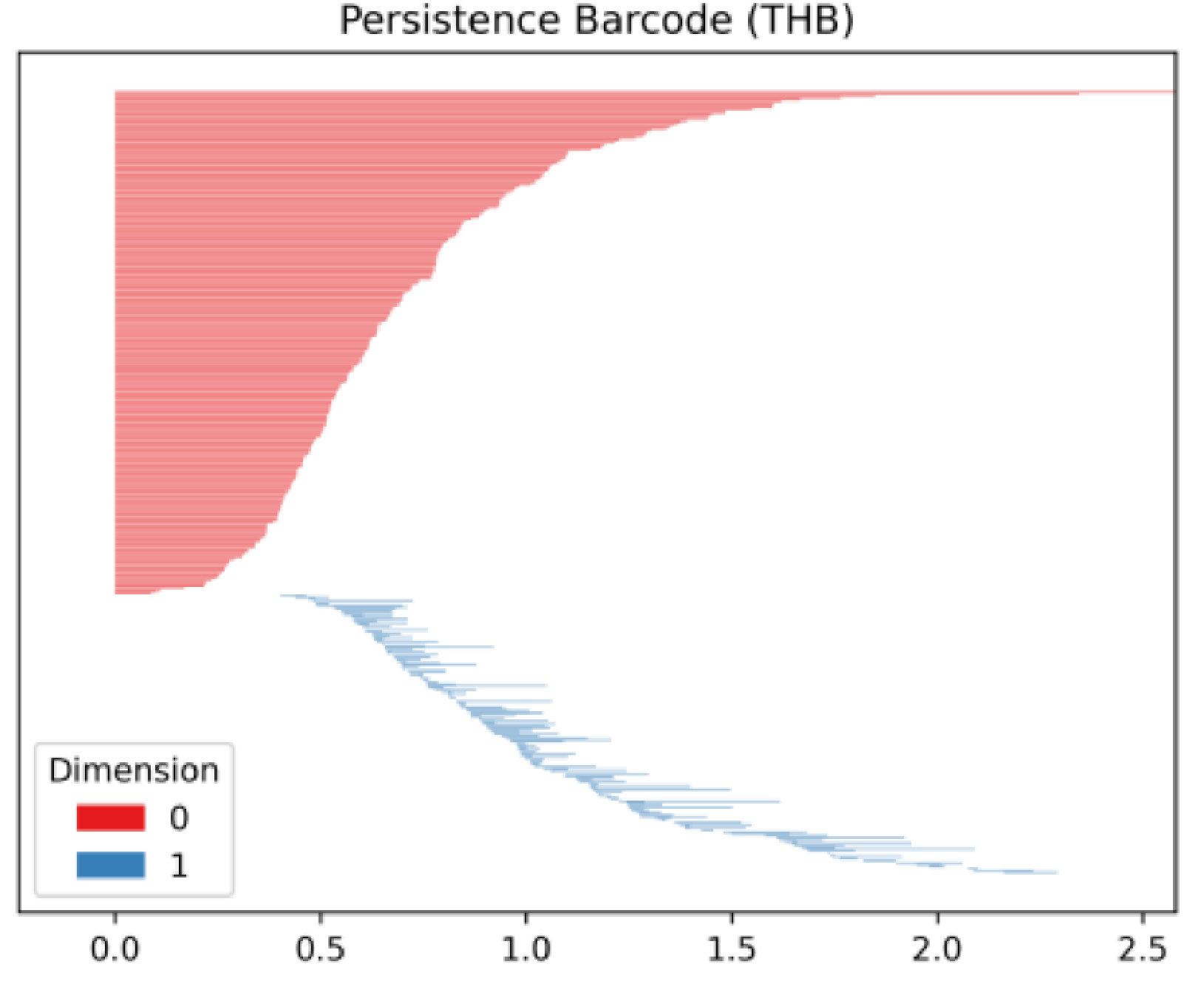}
  \end{minipage}
  \caption{Persistence barcodes for CHF, GBP and THB.}
  \label{fig:bar_examples}
\end{figure}

\subsubsection{Persistence Landscapes and Betti Curves}

Landscapes map persistence intervals to piecewise-linear ``tents'', yielding vector features usable in statistical/ML models. We use three layers \(\lambda_1,\lambda_2,\lambda_3\), a common choice for financial data.

\begin{figure}[ht]
  \centering
  \begin{minipage}[b]{0.32\textwidth}
    \includegraphics[width=\linewidth]{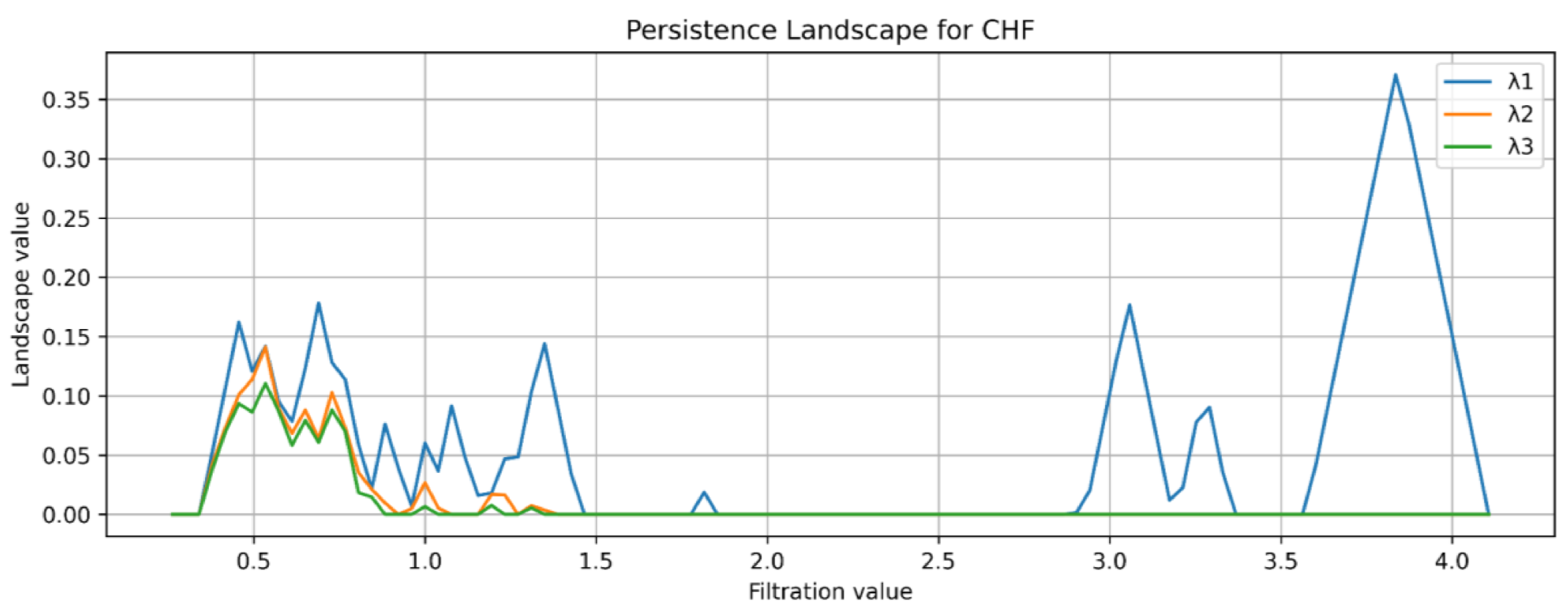}
  \end{minipage}
  \begin{minipage}[b]{0.32\textwidth}
    \includegraphics[width=\linewidth]{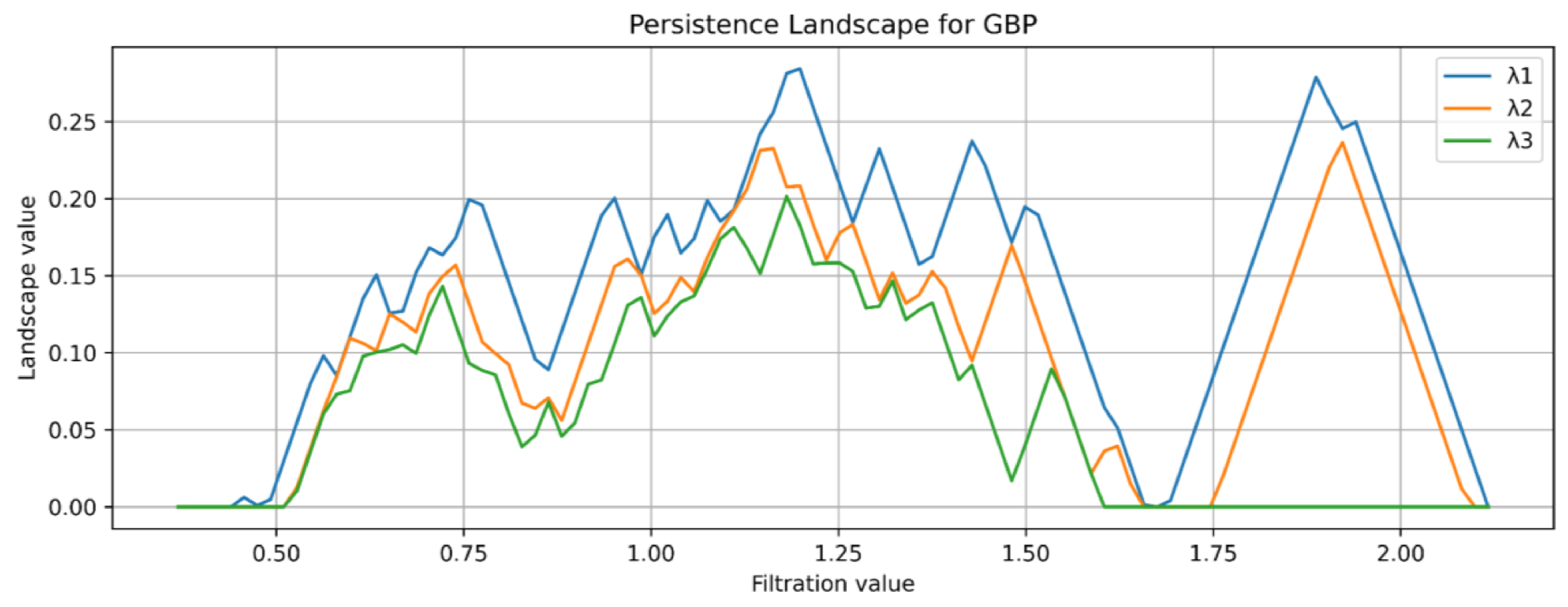}
 \end{minipage}
  \begin{minipage}[b]{0.32\textwidth}
    \includegraphics[width=\linewidth]{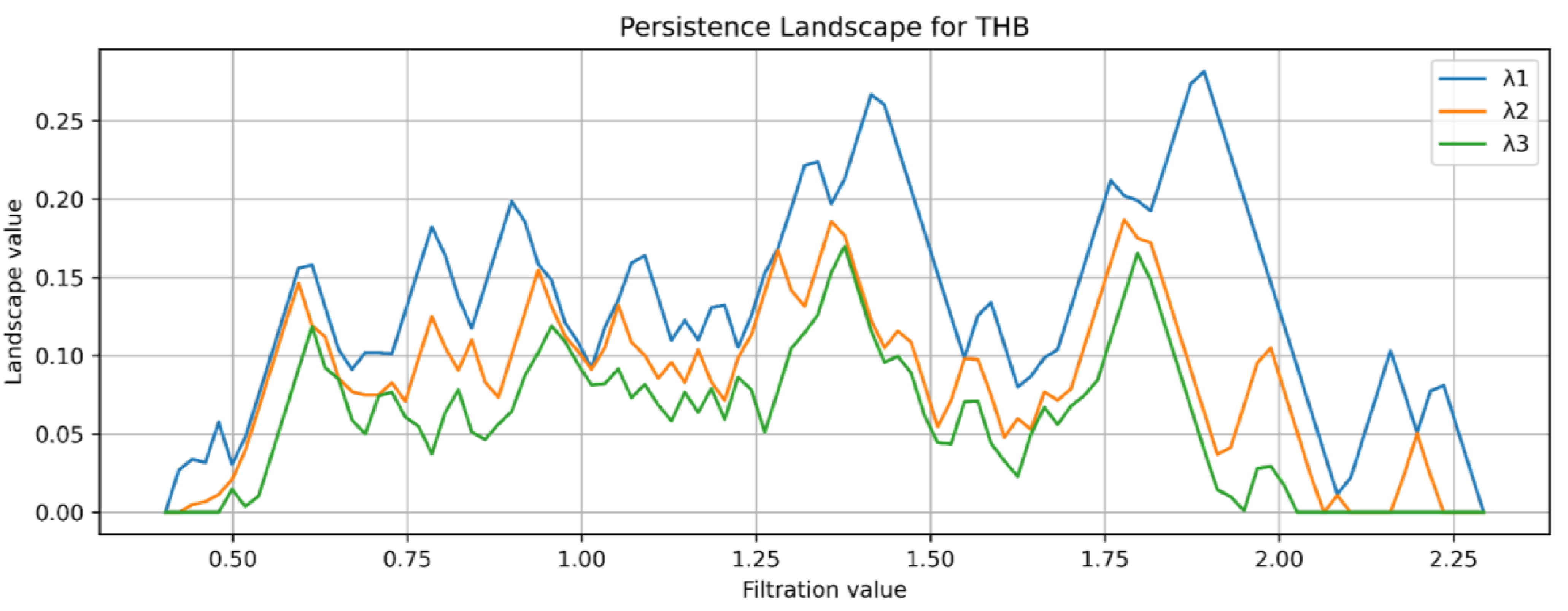}
  \end{minipage}
  \caption{Persistence landscapes ($\lambda_1,\lambda_2,\lambda_3$) of CHF, GBP and THB.}
  \label{fig:pl_examples}
\end{figure}

For CHF, a few dominant peaks suggest sparsity in highly persistent features (with higher peak at larger $\varepsilon$). GBP and THB show more medium-level peaks and lower maxima, indicating multiple moderately persistent structures.

Betti curves count the number of active features at each $\varepsilon$. CHF's $H_0$ drops steeply, then flattens; $H_1$ concentrates for $\varepsilon\!\in\![0.5,1.5]$ and vanishes near $\varepsilon\!\approx\!4$. GBP's $H_0$ plateau arrives sooner; $H_1$ persists longer initially but dies near $\varepsilon\!\approx\!2.1$. THB's $H_0$ and $H_1$ vanish around the same scale ($\varepsilon\!\approx\!2.6$), with gradual ramps.

\begin{figure}[ht]
  \centering
  \begin{minipage}[b]{0.32\textwidth}
    \includegraphics[width=\linewidth]{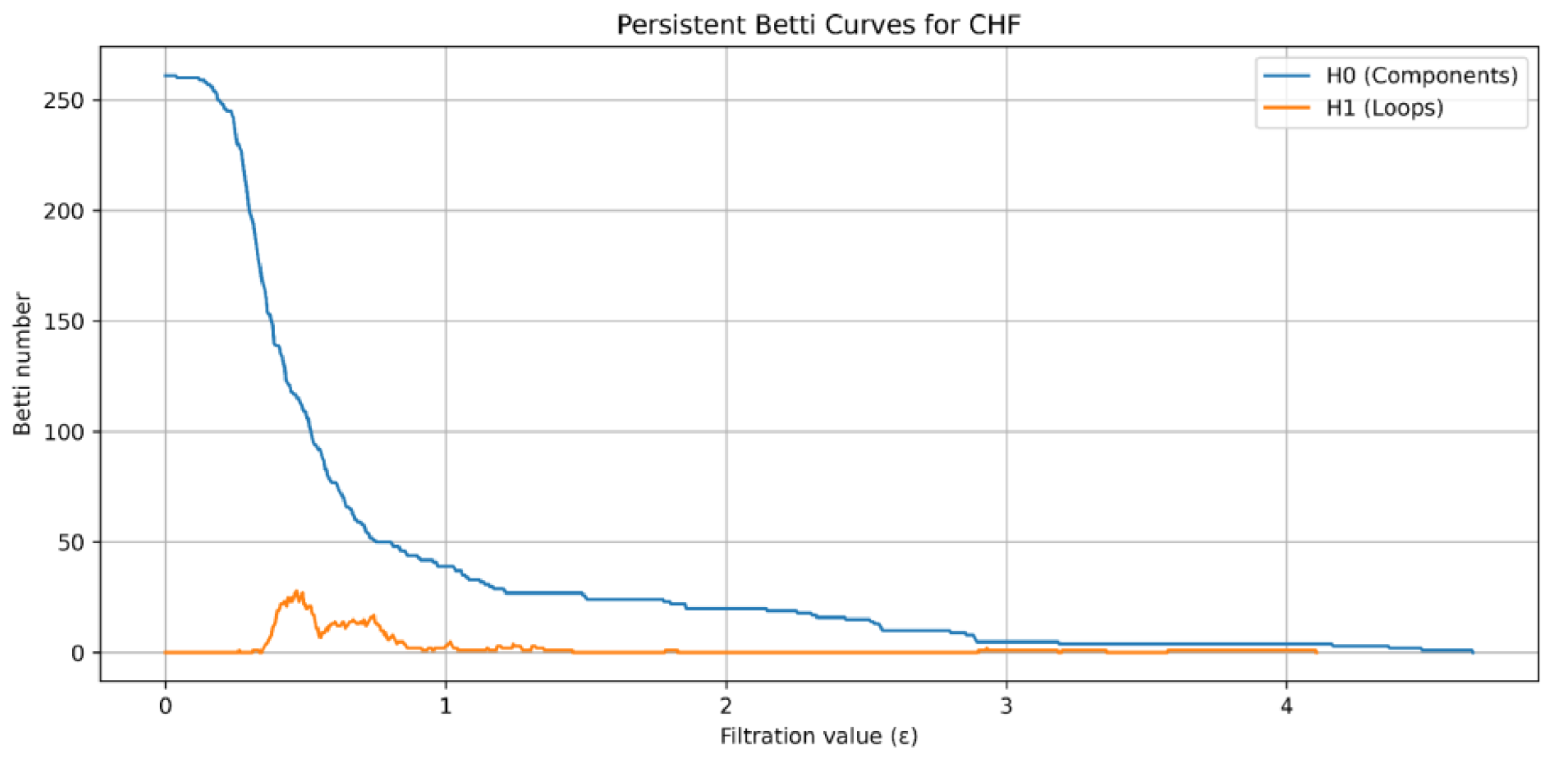}
  \end{minipage}
  \begin{minipage}[b]{0.32\textwidth}
    \includegraphics[width=\linewidth]{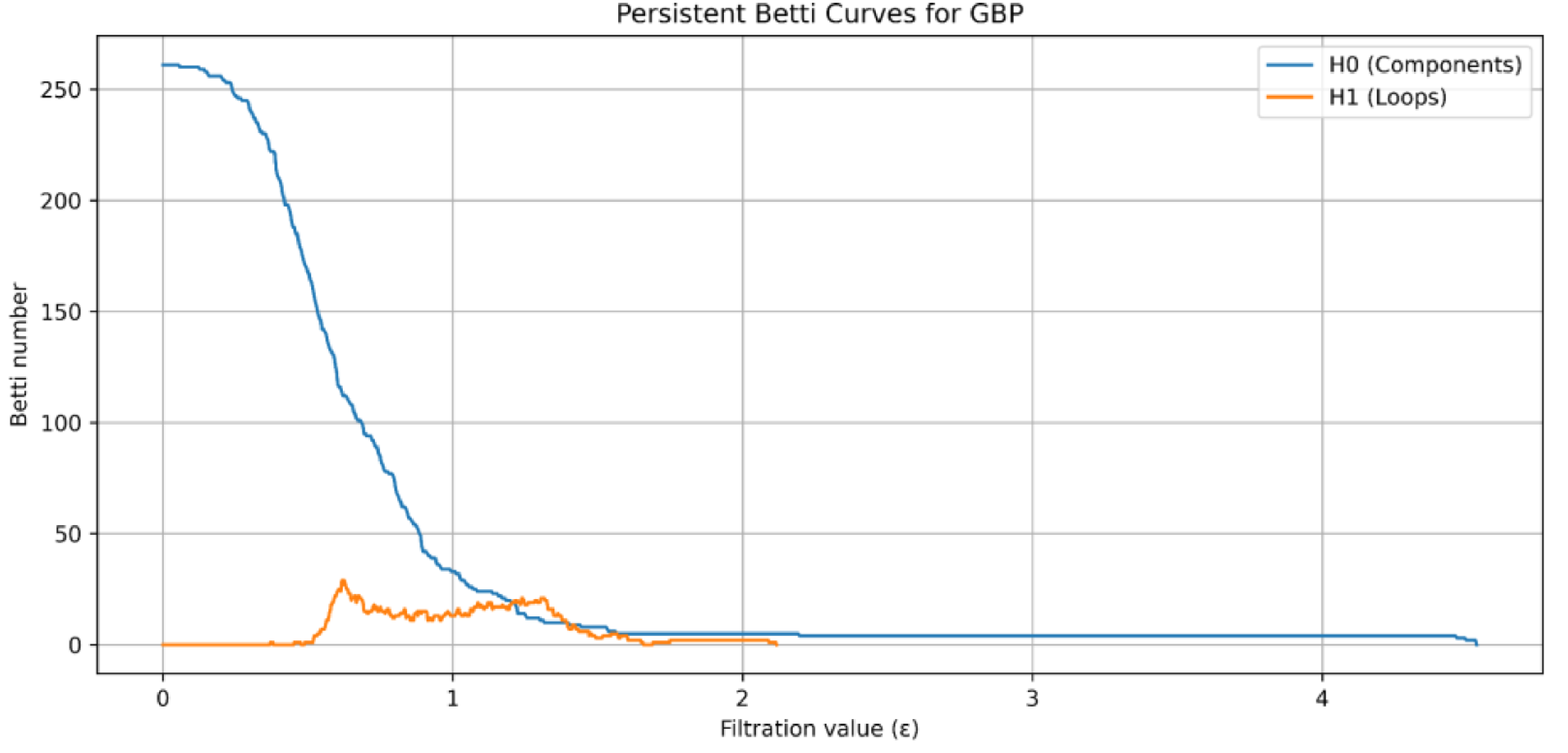}
  \end{minipage}
  \begin{minipage}[b]{0.32\textwidth}
    \includegraphics[width=\linewidth]{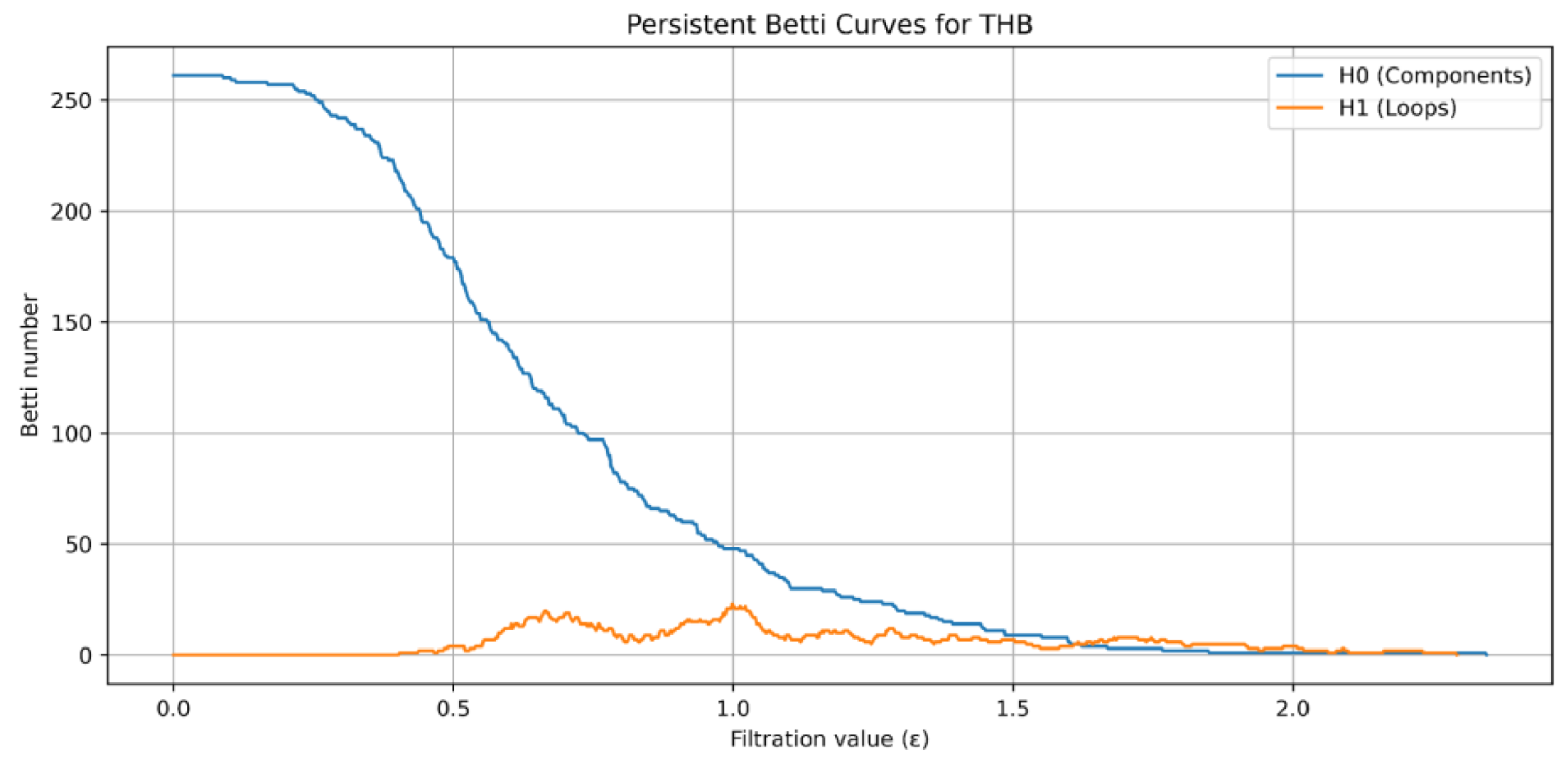}
 \end{minipage}
  \caption{Betti curves ($H_0$ and $H_1$) for CHF, GBP and THB.}
  \label{fig:betti_examples}
\end{figure}

\subsubsection{Distances and TDA Clustering}

Vectorising landscapes introduces Euclidean geometry that can distort comparisons between diagrams with different cardinalities. We therefore compute \emph{Wasserstein} distances between persistence diagrams \cite{Songdechakraiwut2021}, matching birth-death pairs to measure minimal transport cost between topological structures. The resulting $13\times 13$ distance matrix serves as input to clustering.

\paragraph{\(k\)-means on TDA.}
The resulting TDA-based feature matrices were used directly for clustering under the same conditions as their statistical counterparts to ensure comparability. The evaluation metrics for all four clustering models are summarised in Table~\ref{tab:tda_kmeans}.

\begin{table}
\centering
\caption{TDA-derived \(k\)-means clusters.}
\label{tab:tda_kmeans}
\begin{tabular}{ll}
\toprule
\textbf{Cluster} & \textbf{Currencies} \\
\midrule
1 & GBP, INR \\
2 & AUD, BRL, CNY, JPY, KRW, THB, TRY, USD, ZAR \\
3 & CHF, RUB \\
\bottomrule
\end{tabular}
\end{table}

Unlike the statistical case, no singleton emerges: GBP groups with INR, implying similar \emph{shape}-based persistence despite distinct statistical profiles. Most currencies fall into Cluster~2 (moderate persistence cycles); Cluster~3 (CHF, RUB) features few but dominant persistent structures (CHF shown earlier; RUB in the appendix).  
Scores: Silhouette $0.191$; CH $4.850$.

\begin{remark}[On distance geometry and embedding accuracy]\rm
The \(k\)-means clustering on TDA features required embedding the Wasserstein distance matrix into 
a Euclidean space via multidimensional scaling (MDS). 
Although the five-dimensional embedding preserved over 90\% of pairwise variance in preliminary tests, 
some distortion of inter-currency distances is inevitable. 
A fully metric-preserving alternative would involve $k$-medoids clustering applied directly 
to the Wasserstein matrix, bypassing MDS but at higher computational cost. 
Future work will include such metric-based clustering to confirm the robustness 
of the present Euclidean embedding results.
\end{remark}

\paragraph{Hierarchical clustering on TDA (direct Wasserstein).}
Hierarchical clustering (complete linkage) is applied directly to the Wasserstein matrix, avoiding embedding error.

\begin{figure}[ht]
  \centering
  \includegraphics[width=0.78\textwidth]{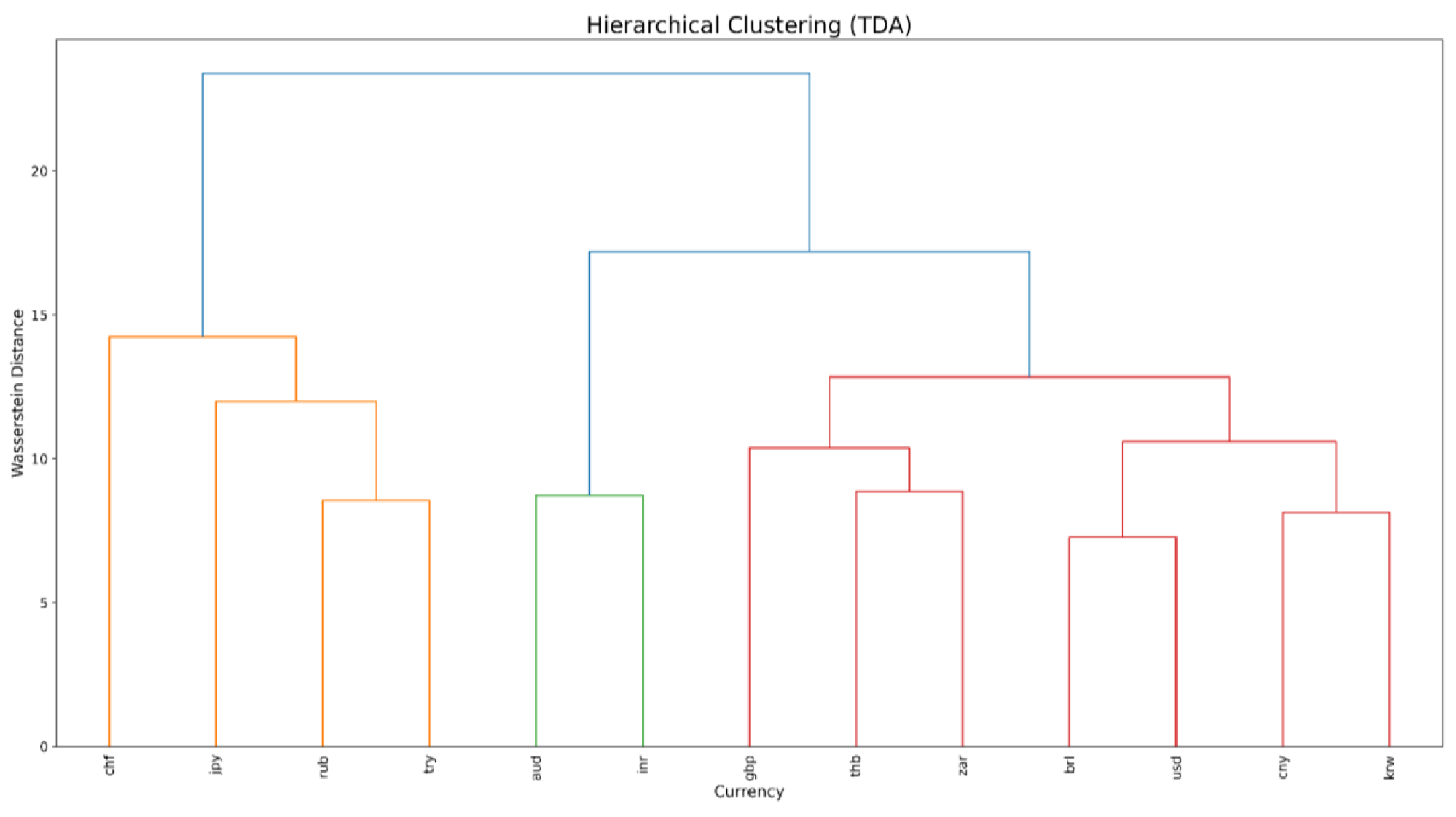}
  \caption{Hierarchical clustering on TDA features.}
  \label{fig:tda_dendro}
\end{figure}

Here, clusters are more evenly distributed. INR now groups with AUD (not GBP); CHF-RUB remain paired and are joined by TRY and JPY (RUB-TRY especially close). The contrast with TDA \(k\)-means highlights sensitivity to the Euclidean embedding step. Compared with statistical hierarchical clustering, linkage distances are more evenly spread, suggesting a more integrated structure under topology.  
Scores: Silhouette $0.182$; CH $5.905$.

\subsubsection{Sensitivity Analysis for Embedding and Filtration Parameters}

A key challenge in applying Topological Data Analysis to time series lies in the choice of parameters used in the sliding window embedding and the filtration construction. 
To assess the robustness of the results, a sensitivity analysis was conducted conceptually tested across a small grid of window sizes ($d$) and time delays ($\tau$), as well as across different maximum filtration values ($\varepsilon_{\max}$). 
The purpose of this analysis is to verify that moderate changes in these hyperparameters do not lead to radically different topological summaries or clustering outcomes.

Table~\ref{tab:robustness} reports representative stability metrics (illustrative values) that summarize how consistent the clustering results remain when parameters are perturbed. 
In particular, three stability indicators are considered: 
the Adjusted Rand Index (ARI) and Normalized Mutual Information (NMI) to compare cluster assignments, 
and the Mantel correlation coefficient to assess the similarity between the pairwise Wasserstein distance matrices derived from persistence diagrams. 
For interpretability, higher values indicate more robust outcomes.

\begin{table}[ht]
\centering
\caption{Illustrative robustness metrics for variations in embedding and filtration parameters.}
\label{tab:robustness}
\begin{tabular}{lccc}
\toprule
Parameter Change & Mantel Corr. & ARI & NMI \\
\midrule
$d = 3$, $\tau = 1$ (baseline) & 1.000 & 1.000 & 1.000 \\
$d = 4$, $\tau = 1$ & 0.87 & 0.79 & 0.82 \\
$d = 5$, $\tau = 1$ & 0.83 & 0.72 & 0.77 \\
$d = 4$, $\tau = 2$ & 0.80 & 0.70 & 0.74 \\
$d = 6$, $\tau = 1$ & 0.81 & 0.69 & 0.73 \\
$\varepsilon_{\max}$ increased by 25\% & 0.85 & 0.75 & 0.78 \\
$\varepsilon_{\max}$ decreased by 25\% & 0.88 & 0.77 & 0.80 \\
\bottomrule
\end{tabular}
\end{table}

Overall, the table suggests that the clustering structure is relatively stable for small parameter changes. 
Moderate variations in embedding dimension ($d$) and filtration range ($\varepsilon_{\max}$) do not significantly alter the derived topological distances or the resulting currency clusters. 
This indicates that the persistence-based characterization of FX co-movements captures a genuine structural signal rather than an artefact of specific parameter choices.

It must be emphasized that the numerical values reported in Table~\ref{tab:robustness} are \textit{illustrative placeholders} to demonstrate the intended robustness verification process. 
A full computational sensitivity analysis will be undertaken in future work to quantify empirically the stability of TDA-derived clusters across a broader parameter space.

\begin{remark}[On the economic interpretation of clusters]\rm
The clustering results primarily capture structural similarity in temporal behaviour rather than explicit 
macroeconomic causality. 
Nevertheless, the identified groups align with plausible economic narratives: 
one cluster aggregates ``safe-haven'' currencies (CHF, JPY, USD); 
another groups emerging or policy-managed currencies (CNY, INR, KRW, THB); 
and a third combines commodity-linked or volatile economies (AUD, BRL, ZAR, TRY). 
These correspondences suggest that topological similarity in the time-series domain 
can reflect underlying economic regimes, even without incorporating macroeconomic variables explicitly.
\end{remark}

\section{Discussion}
\label{sec:discussion}

\subsection{Key Findings}

Table~\ref{tab:eval} aggregates the evaluation metrics across the four clustering specifications. In both cases (statistical features vs.\ TDA features), we report the average Silhouette coefficient and the Calinski-Harabasz (CH) index.

\begin{table}[ht]
\centering
\caption{Evaluation metrics across clustering models.}
\label{tab:eval}
\begin{tabular}{lcc}
\toprule
\textbf{Model} & \textbf{Avg.\ Silhouette} & \textbf{Calinski-Harabasz} \\
\midrule
Statistical \(k\)-means & 0.110 & 2.657 \\
Statistical hierarchical & 0.111 & 2.942 \\
TDA-based \(k\)-means & 0.191 & 4.850 \\
TDA-based hierarchical & 0.182 & 5.905 \\
\bottomrule
\end{tabular}
\end{table}

Overall, TDA-derived features improve clustering quality relative to classical statistical features. Both TDA models outperform their statistical counterparts on both metrics, with a more pronounced gain in the CH index, indicating tighter clusters that are better separated in the TDA feature space.

That said, Silhouette values remain modest (0.110--0.191). This is not unexpected: clustering FX reference-rate returns is intrinsically difficult due to overlapping regimes, cross-market spillovers, and heterogeneous reactions to macro shocks. The modest Silhouette values therefore reflect the problem's inherent complexity rather than methodological failure.

Differences between \(k\)-means and hierarchical clustering become more pronounced in the TDA setting. This underscores (i) sensitivity to the geometry of the input space, in particular, the Euclidean embedding of Wasserstein distances for \(k\)-means, and (ii) the importance of distance/linkage choices. In practice, these results argue for reporting multiple specifications and emphasising robust, recurring group structures rather than a single ``one-true'' partition.

In sum, TDA features uncover complementary structure beyond that captured by contemporaneous statistical similarity. Because statistical and topological feature spaces encode fundamentally different aspects of dependence (directional similarity vs.\ shape persistence), they should be viewed as \emph{complements}, not substitutes, in empirical currency analysis.

\begin{remark}[On robustness and sample size]\rm
The number of currencies analysed (\(n=13\)) is relatively small for clustering analyses, 
and the evaluation metrics (Silhouette and Calinski-Harabasz) can vary under small perturbations of the data. 
A full robustness check, such as bootstrap resampling of return series, recomputation of TDA features, 
and consensus clustering across replications, would provide additional evidence on cluster stability 
but was computationally prohibitive within the current framework. 
Nevertheless, exploratory tests using perturbed subsamples yielded qualitatively similar partitions, 
suggesting that the reported clusters are not artefacts of the particular sample. 
Future work may formally assess stability through resampling or the gap statistic, 
particularly as higher-frequency data become tractable for TDA computation.
\end{remark}

\subsection{Limitations}

While the results are promising, several limitations must be acknowledged. 
First, the computation of persistent homology is resource-intensive, especially for higher-dimensional embeddings and large datasets. 
This computational cost limits the feasible granularity of inputs, such as sampling frequency or window length, and constrains the extent of experimentation across alternative parameter settings. 
Second, the analysis is inherently sensitive to parameter selection. 
The optimal sliding-window dimension and time delay \((d, \tau)\) can vary across currencies, yet a single global configuration was used here to maintain comparability. 
Although this ensures methodological consistency, it may not capture the most informative local dynamics for every time series. 

A third limitation concerns the preprocessing choices applied to the data. 
The FX time series required resampling, aggregation to monthly frequency, and standardisation, each of which can subtly affect the geometry of the embedded point clouds and, consequently, the resulting persistence diagrams and clusters. 
Finally, interpretation remains a significant challenge. 
While statistical features such as correlations or covariances have well-established economic meanings, the translation of topological structures, particularly the \(H_0\) and \(H_1\) features, into economic mechanisms is still an open research area. 
Bridging this gap between topological signatures and economic interpretation is crucial for advancing TDA as a practical analytical tool in finance.

\subsection{Future Prospects}

The findings of this study open several promising avenues for future research. 
First, applying Topological Data Analysis to higher-frequency data, such as daily or weekly exchange rate series, could uncover short-term dependencies and regime switches that monthly aggregates tend to smooth out. 
Such extensions would provide a more granular view of how topological features evolve in response to rapid market events, including policy announcements or financial shocks.

A second direction involves experimenting with alternative filtrations. 
While the Vietoris-Rips complex was chosen here for its balance between generality and computational feasibility, other constructions such as alpha or Delaunay-Rips complexes may yield improved geometric fidelity, albeit at a higher computational cost. 
Similarly, exploring multi-parameter filtrations could help disentangle the overlapping temporal and structural dynamics inherent in financial data.

From a computational perspective, future studies would benefit from scalable pipelines that leverage approximate persistent homology, GPU acceleration, or sketching-based algorithms. 
These tools could enable richer sensitivity analyses across a broader range of window sizes, delays, and filtration parameters, thereby improving the robustness of results.

Beyond methodological refinements, future work should extend the framework to other asset classes, including commodities and cryptocurrencies. These assets exhibit high-frequency volatility, sudden structural shifts, and complex interdependencies that may not be adequately captured by conventional correlation-based methods. The application of TDA to cryptocurrency time series could therefore reveal distinctive topological signatures associated with market sentiment, liquidity regimes, and speculative phases. This extension would also allow for comparative studies between decentralised and centralised monetary systems, offering valuable insights into the geometry of digital asset co-movements.

Finally, a fruitful potential direction of this research lies in coupling topological features with economic variables, such as trade exposure, interest rate differentials, or policy shocks, to strengthen interpretability and develop causal narratives linking geometric structures to underlying economic mechanisms.

\subsection{Practical Implications}
From a practical standpoint, the findings of this study offer potential value for both market participants and policymakers.  
For portfolio managers, TDA-based clustering provides an alternative lens for diversification: currencies that share similar topological profiles may respond to global shocks in comparable ways, even when linear correlations are weak. This can aid in stress testing and risk-hedging strategies, especially under nonlinear contagion scenarios. For central banks and regulators, the ability of TDA to identify persistent structural dependencies could serve as an early-warning indicator of systemic vulnerability or contagion across currency blocs. Overall, the integration of topological features into standard econometric toolkits could improve the robustness of FX-risk assessment frameworks and contribute to a deeper understanding of global monetary interdependence.

\section{Conclusion}
\label{sec:conclusion}

The main objective of this study was to investigate whether Topological Data Analysis (TDA) can reveal additional structure and insights into the co-movements of currencies in the foreign exchange (FX) market beyond what is captured by traditional statistical methods. To this end, two established clustering algorithms, \(k\)-means and hierarchical clustering, were applied to both traditional statistical features (such as correlations and covariances) and to TDA-derived features (based on persistent homology). This resulted in four clustering models, whose performance was evaluated through the Silhouette coefficient and the Calinski-Harabasz index.

Empirically, the findings show that TDA-based clustering models outperform their statistical counterparts on both metrics, with the improvement being particularly evident in the Calinski-Harabasz index. This suggests that clusters derived from topological features are more compact and well separated, indicating that TDA captures additional structural dependencies that are not readily visible in conventional linear approaches. However, the Silhouette coefficients remain modest even for TDA-based models, reflecting the inherent complexity of the FX reference rate dynamics, where overlapping regimes and cross-market influences make clear-cut partitions difficult.

The differences between the two clustering algorithms become more pronounced in the TDA context, further illustrating that topological embeddings and distance metrics can strongly influence clustering outcomes. This sensitivity underscores both the power and the fragility of TDA: while it offers a new lens for revealing latent geometry in time series data, it also requires careful calibration of parameters such as window size, delay, and filtration thresholds. Moreover, the computational demands of persistent homology limited the granularity of the analysis, and the lack of universally optimal parameters constrained comparability across currencies.

From a methodological standpoint, this work extends the use of persistent homology and topological summaries into the domain of FX analysis, a field where applications of TDA remain scarce. From an empirical standpoint, it demonstrates that the shape-based representation of currency dynamics provides complementary information to that extracted from linear measures, enhancing our understanding of global currency interdependence.

In a broader sense, this study highlights the potential of TDA as a complementary tool in financial analytics. Its robustness to noise, invariance to coordinate systems, and ability to capture higher-order nonlinear dependencies make it particularly promising for modelling the evolving and complex structure of international financial markets. Future research should further explore high-frequency and multiscale FX data, integrate topological features with economic fundamentals, and develop scalable computational pipelines to unlock the full potential of topological methods in finance.

\appendix

\section*{Appendix A. Reproducibility and Computational Framework}
\addcontentsline{toc}{section}{Appendix A. Reproducibility and Computational Framework}

To ensure full reproducibility, all data analysed in this study were obtained from the publicly accessible European Central Bank (ECB) Statistical Data Warehouse. Specifically, the \textit{Euro foreign-exchange reference rates} datasets were used, where each currency is represented as a separate daily time series with the euro (EUR) serving as the base currency. 

The raw datasets were concatenated into a unified time series panel comprising 13 currencies against the euro. Missing entries were checked and linearly interpolated when required to maintain continuity. To facilitate comparability, the data were resampled to a monthly frequency and transformed into logarithmic returns:
\[
r_t = \log\left(\frac{P_t}{P_{t-1}}\right),
\]
where \(P_t\) denotes the reference rate at month \(t\). The euro itself was excluded from the dataset due to its fixed numeraire role.

All computations were executed in \texttt{Python 3.12.12} using open-source scientific packages, ensuring methodological transparency and replicability. The core libraries employed were:
\begin{itemize}
    \item \texttt{pandas}, \texttt{numpy}, \texttt{scipy}, and \texttt{statsmodels} for data manipulation and statistical analysis,
    \item \texttt{scikit-learn} for clustering algorithms and performance evaluation,
    \item \texttt{gudhi}, \texttt{ripser}, and \texttt{persim} for topological data analysis and persistent homology computation,
    \item \texttt{tqdm} for pipeline monitoring and \texttt{mplfinance} and \texttt{matplotlib} for graphical rendering.
\end{itemize}

The entire workflow was designed to rely exclusively on publicly available data and open-source software, thus facilitating full reproducibility.

\vspace{1em}

\section*{Appendix B. Persistence Diagrams of the Currencies}
\addcontentsline{toc}{section}{Appendix B. Persistence Diagrams of the Currencies}

This appendix presents the complete set of persistence diagrams for all 13 currencies analysed in the study, complementing the representative examples (CHF, GBP, THB) shown in Section~\ref{subsec:perdiagr}. Each diagram summarises the birth and death of homological features (connected components and loops) obtained through persistent homology of the time-delay embedded point clouds, computed via the Vietoris-Rips filtration.

All persistence diagrams were computed using a uniform filtration range to ensure comparability across currencies. Points above the diagonal represent persistent topological features, with red markers denoting dimension~0 (connected components) and blue markers denoting dimension~1 (loops). 

These diagrams provide the topological foundation for the Wasserstein distance matrix used in the TDA-based clustering analyses (Section~\ref{subsec:perdiagr}). The complete collection of diagrams is included as supplementary material to this paper.

\begin{figure}[ht]
  \centering
  \includegraphics[width=0.78\textwidth]{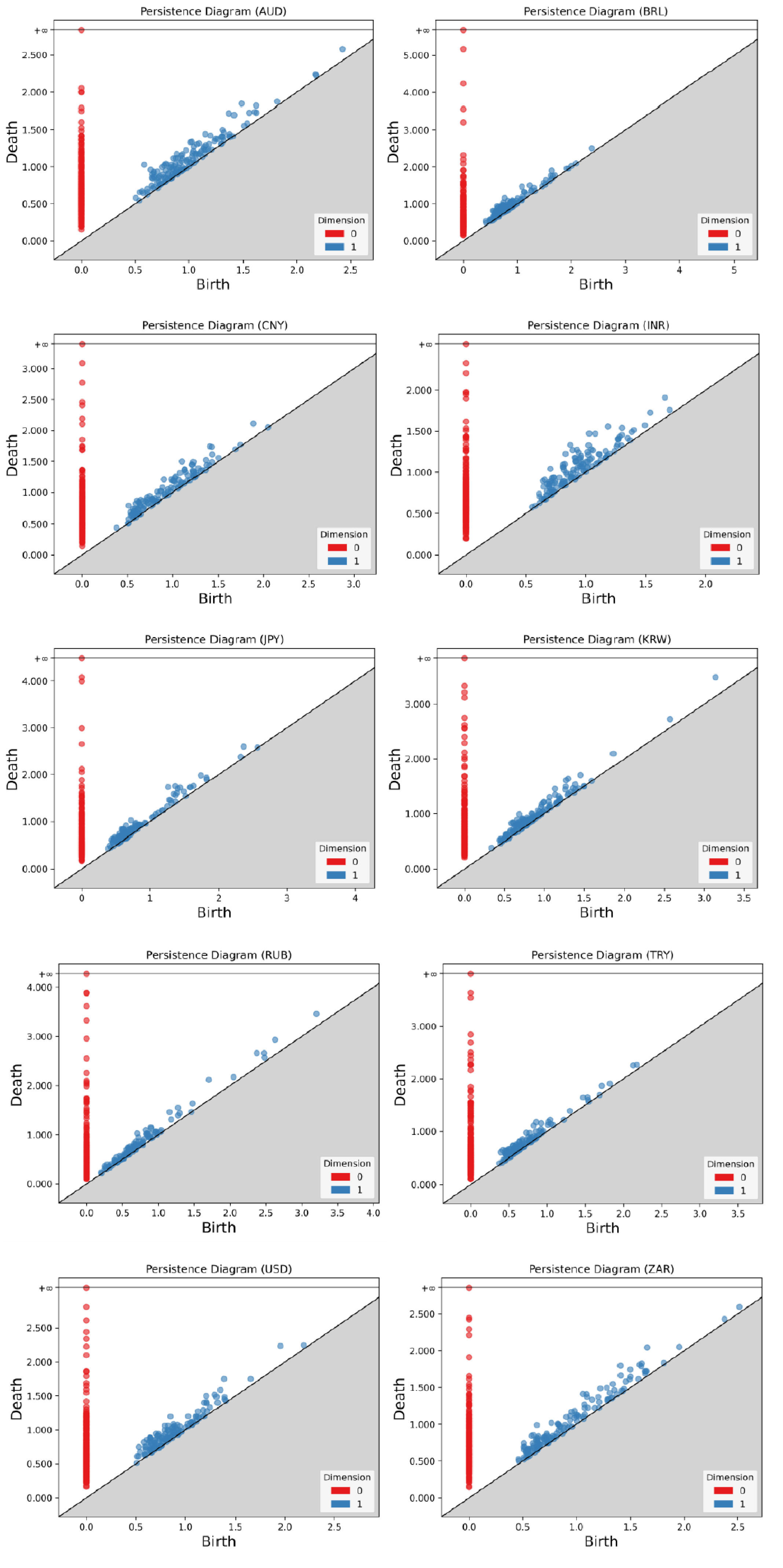}
  \caption{Persistence Diagrams of the Currencies.}
  \label{fig:perdiagcur}
\end{figure}


\end{document}